\documentclass{article}
\usepackage{leaplab}
\usepackage{placeins}
\usepackage[table,x11names]{xcolor}
\usepackage[utf8]{inputenc}
\usepackage[T1]{fontenc}
\usepackage{url}
\usepackage{booktabs}
\usepackage{amsfonts}
\usepackage{amsmath}
\usepackage{amssymb}
\usepackage{amsthm}
\usepackage{bm}
\usepackage{nicefrac}
\usepackage{wrapfig}
\usepackage{microtype}
\usepackage{graphicx}
\usepackage{caption}
\usepackage{algorithm}
\usepackage{algorithmic}
\usepackage{multirow}
\usepackage[most]{tcolorbox}
\usepackage{enumerate}
\usepackage{pifont}
\usepackage{fontawesome5}
\usepackage{enumitem}
\usepackage{marvosym}
\usepackage[colorlinks,linkcolor=red,citecolor=blue,urlcolor=blue]{hyperref}

\newtheorem{theorem}{Theorem}[section]

\newtheorem{definition}[theorem]{Definition}
\theoremstyle{remark}

\definecolor{lightgrey}{RGB}{247, 247, 247}
\newenvironment{leapabstract}{
  \begin{tcolorbox}[
    colback=lightgrey,
    colframe=white,
    boxrule=0pt,
    arc=10pt,
    left=16pt,
    right=16pt,
    top=12pt,
    bottom=12pt,
    width=\textwidth,
    enlarge left by=0mm,
    before skip=10pt,
    after skip=10pt
  ]
}{
  \end{tcolorbox}
}

\makeatletter
\def\icmldate#1{\gdef\@icmldate{#1}}
\icmldate{\today}
\makeatother

\makeatletter
\fancypagestyle{fancytitlepage}{
  \fancyhead{}
  \lhead{\includegraphics[height=1.5cm]{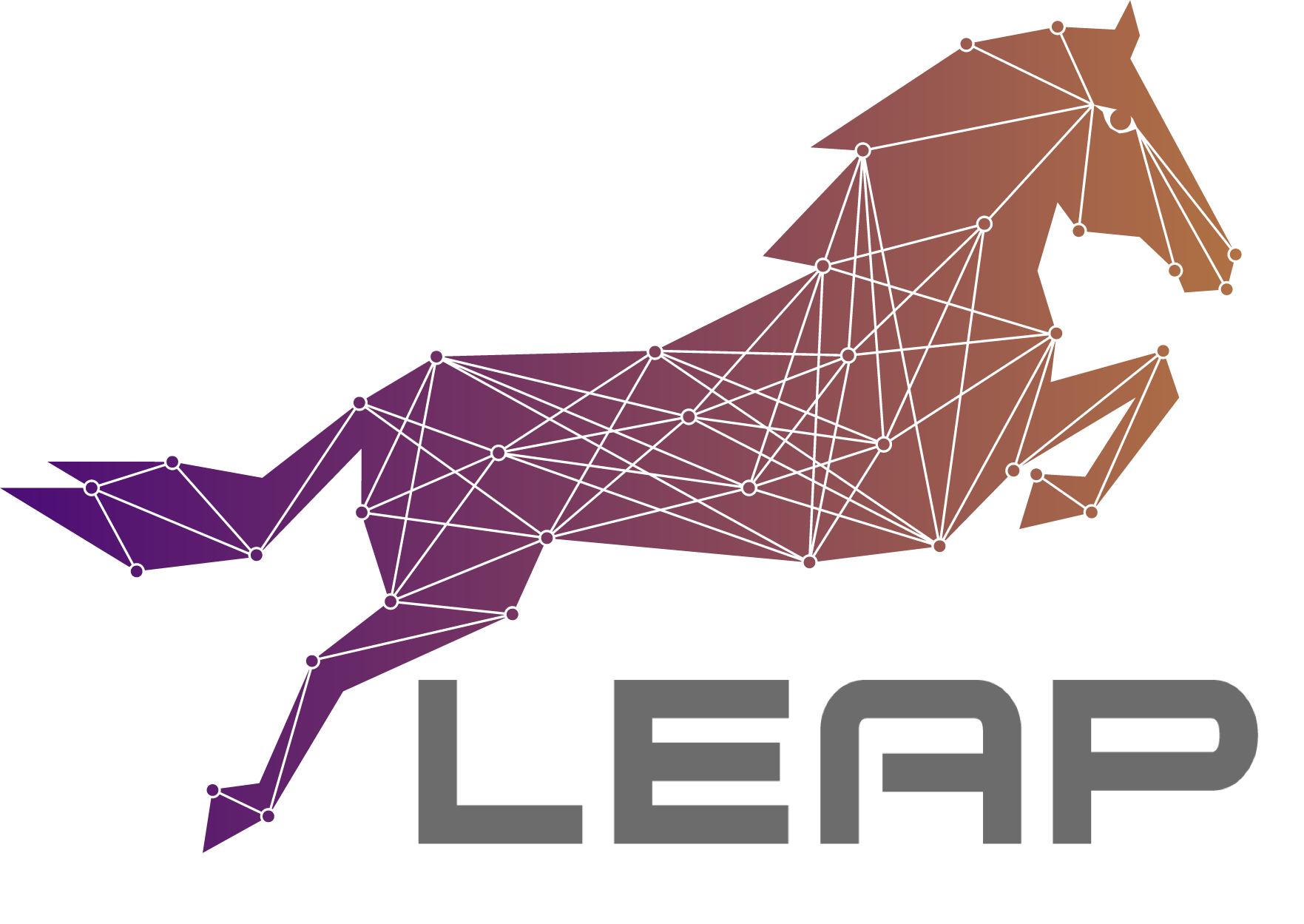}}
  \rhead{\it \@icmldate}
  \cfoot{}
}
\makeatother

\icmltitlerunning{From Reasoning Chains to Verifiable Subproblems: SCRL Enables Credit Assignment for LLM Reasoning}

\begin{document}

\icmltitle{From Reasoning Chains to Verifiable Subproblems: Curriculum Reinforcement Learning Enables Credit Assignment for LLM Reasoning}

\begin{icmlauthorlist}
\mbox{Xitai Jiang$^{\,1,2\,*\,\dagger}$},
\mbox{Zihan Tang$^{\,2\,*}$},
\mbox{Wenze Lin$^{\,1,2\,*}$},
\mbox{Yang Yue$^{\,1}$},
\mbox{Shenzhi Wang$^{\,1}$},
and \mbox{Gao Huang$^{\,1\,\textrm{\Letter}}$}
\end{icmlauthorlist}

$^{1\,}$LeapLab, Tsinghua University \quad
$^{2\,}$Qiuzhen College, Tsinghua University

$^{*}$ Equal Contribution \quad $^{\dagger}$ Project Lead \quad $^{\textrm{\Letter}}$ Corresponding Author

\icmlcorrespondingauthor{\{jiang-xt21,tangzh23,linwz25\}@mails.tsinghua.edu.cn, gaohuang@tsinghua.edu.cn}{}

\vskip .3in

\printNotice{}

\begin{leapabstract}
Reinforcement learning from verifiable rewards (RLVR) has shown strong promise
  for LLM reasoning, but typical outcome-based RLVR methods remain inefficient on
  hard problems. Correct final-answer rollouts are rare, and standard
  sample-level credit assignment fails to leverage partial reasoning progress
  embedded in unsuccessful attempts.
To address this problem, we introduce \textbf{SCRL} (\textbf{S}ubproblem \textbf{C}urriculum \textbf{R}einforcement
  \textbf{L}earning), a curriculum reinforcement learning framework built on verifiable
  subproblems derived from reasoning chains. Given a reference solution, SCRL derives a series of verifiable subproblems and constructs a subproblem curriculum, with the final subproblem
  fixed as the original problem. This converts partial progress on hard problems
  into verifiable learning signals. Algorithmically, we propose \emph{subproblem-level normalization}, a training technique based on RLVR
  that normalizes rewards independently at each subproblem position within the rollout group. By assigning the resulting advantages to the corresponding answer
  spans, we enable finer-grained credit assignment without external rubrics or
  reward models.
Our theoretical analysis shows that this subproblem curriculum makes hard problems more learnable by
  lifting them out of gradient dead zones, with larger relative gains as the
  original problem becomes harder. Across seven mathematical reasoning benchmarks, SCRL outperforms strong
  curriculum-learning baselines, yielding +4.1 and +1.9 average-point gains
  compared to GRPO on Qwen3-4B-Base and Qwen3-14B-Base respectively. On three
  hard benchmarks (AIME24, AIME25, and IMO-Bench), SCRL further yields point gains of +3.7 in pass@$1$ and +4.6 in pass@$64$ on
  Qwen3-4B-Base, suggesting improved exploration on hard reasoning problems. Further ablations show that the proposed credit
  assignment is effective and that the gains do not require highly curated
  subproblems or strong external generators, highlighting SCRL as a practical curriculum framework that enables fine-grained credit
  assignment for LLM reasoning.

\end{leapabstract}
\section{Introduction}
\label{sec:intro}

\begin{figure*}[t]
      \centering
      \includegraphics[width=\textwidth]{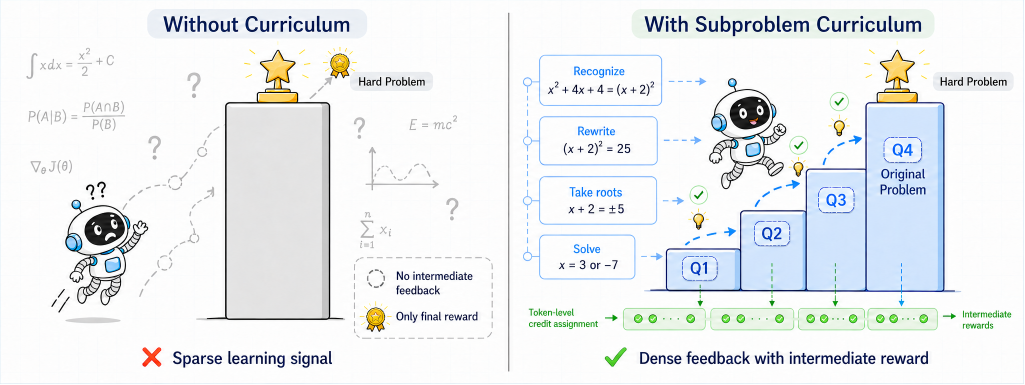}
      \caption{
      Main idea of SCRL. Standard outcome-based RLVR provides only sparse
      final-answer rewards on hard problems. SCRL instead decomposes a hard problem
      into verifiable subproblems, turning intermediate progress into dense learning
      signals and enabling finer-grained credit assignment.
      }
      \label{fig:main_idea}
  \end{figure*}

Reinforcement learning from verifiable rewards~(RLVR) has emerged as a
dominant paradigm for training large language models on mathematical
reasoning, delivering strong empirical gains across benchmarks spanning
grade-school arithmetic to olympiad-level competition~\citep{deepseekr1,Yu2025DAPOAO, DeepSeekMath, jaech2024openai, Wen2025ReinforcementLW}.
The key to its success is that a correct final answer provides an unambiguous and
  automatically checkable reward signal. This removes the need for costly human
  annotation and avoids the reward hacking risks of learned reward
  models~\citep{Skalse2022DefiningAC}.

A central goal of RLVR is to help models solve previously unsolved problems and
  improve their reasoning ability. However, prior work suggests that direct RLVR often
  improves sampling efficiency more than it substantially expands the model's
  capability boundary~\citep{yue2025does,shojaee2025illusion,alam2025limits}. Further studies indicate that training at the edge of the model's current
  capability with challenging problems is key for better reasoning
  ability~\citep{Pikus2025HardEA,li2026questa,dai2026harder,Ma2025LearningWR}.
  This makes hard problems particularly valuable for RL training. Yet typical RLVR methods like GRPO~\citep{deepseekr1} struggle
  precisely on these problems. First, rewards are normalized within a group of
  rollouts sampled from the same prompt, so a group in which all rollouts fail
  provides no learning signal. Second, outcome-based RLVR assigns one \emph{sample-level} advantage to the
  entire rollout. Thus, a near-miss attempt receives the same credit as an
  immediate failure. It is therefore crucial to
  extract learning signals from such hard-but-informative problems.

A natural way to learn from hard problems is to make better use of expert
  trajectories. Existing methods mainly follow two routes. One route is compensating for sparse rewards by training the model to imitate expert-generated trajectories, such as supervised fine-tuning and some off-policy RL methods~\citep{Li2025LLMsCE,Luffy,fu2025srft,zhang2025stephint,lv2025towards}. However, they replace the
  model's own on-policy exploration with supervised imitation, and the resulting
  distribution shift between the expert and student policies can hurt training
  stability and out-of-distribution generalization~\citep{Shenfeld2025RLsRW,chu2025sft}.
  The other route uses on-policy curriculum RL. These methods provide an expert
  reasoning prefix or other hints and train the model to complete the remaining solution.~\citep{Amani2025RLFR,zhang2025scaf,wu2025thought,qiyuan2026hipo,qu2026pope,Luffy,Shi2026R3LRR}. However, these hints are treated as fixed conclusions rather than targets the
  model must derive, so the
  model does not need to discover the critical reasoning steps on its own, and the
  supplied context still shifts the model away from its own generation
  distribution. In fact, solving hard problems requires the model to explore and master the
  intermediate conclusions behind these hints by itself. This raises a central question: \textbf{how can we build a curriculum for hard problems
  that keeps the model exploring on its own, while also properly giving credit to the
  intermediate progress it solves along the way?}
\begin{figure*}[t]
  \centering
  \includegraphics[width=0.98\textwidth]{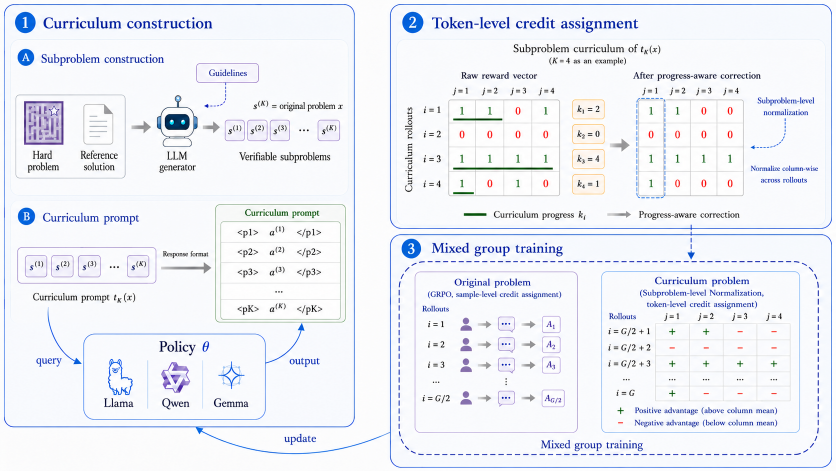}
\caption{
  Overview of SCRL. SCRL constructs verifiable subproblems from a reference
  solution, uses structured responses to assign subproblem-level rewards back to
  answer-span tokens, and jointly trains curriculum rollouts with original-problem
  rollouts through mixed group training.
  }
  \label{fig:colwise_grpo}
\end{figure*}

We propose \textbf{SCRL}~(\textbf{Subproblem Curriculum Reinforcement Learning}),
  drawing inspiration from a familiar structure in mathematical competitions: the
  multi-part problem.
  In a competition exam, a hard problem is broken into a sequence of subproblems
  of increasing difficulty, all visible at once; solving an earlier part yields a
  result that serves as a natural basis for the next.
  Given the expert solution to a hard problem, we offline construct a sequence of
  $K$ verifiable subproblems using an external LLM. The subproblems are ordered from easier to harder, with each later subproblem
  building on the previous ones, and each subproblem has a verifiable answer. We fix the final subproblem as the original problem itself and ask the model to
  answer all $K$ subproblems in a \emph{single on-policy rollout}.
  This organically realizes a curriculum learning structure: when the model
  correctly solves an earlier subproblem, its answer becomes a natural basis
  for the next, guiding the model toward increasingly difficult reasoning.
  Critically, the reasoning steps that bridge consecutive subproblems are
  self-produced, earned through the model's own on-policy rollout. These intermediate results
  provide verifiable process-level supervision, naturally enabling finer-grained
  credit assignment within the rollout.
  We realize this through \emph{subproblem-level normalization}, a novel RLVR
  training technique that normalizes rewards independently at each subproblem position
  and assigns the resulting advantages to the corresponding answer spans. In particular, to prevent the model from rewarding later subproblems
  without solving earlier ones, we align credit 
   with curriculum progress by counting
  only the longest consecutively solved subproblem sequence.
For example, the subproblem reward $[1,1,0,1]$ is treated as $[1,1,0,0]$,
  because progress after the first failed subproblem is not credited.

We validate SCRL with both theory and experiments. Theoretically, we show that
  subproblem decomposition lifts hard problems out of gradient dead zones by
  recovering non-degenerate learning signals from earlier subproblems. We formalize
  this as a metric recovery result, where optimization is lifted from the original
  policy manifold to a subproblem product manifold and the recovery ratio grows
  with problem difficulty. The empirical results are consistent with this prediction: SCRL improves over
  strong curriculum-learning baselines across mathematical reasoning benchmarks.
  Ablations further confirm the effectiveness of subproblem-level credit assignment
  and show that SCRL does not rely on highly curated subproblems or strong
  subproblem generators.
  
Our main contributions are:
\begin{itemize}
    \item \textbf{SCRL framework for curriculum learning.}
      We propose a curriculum RL framework that turns each hard problem into a
      sequence of verifiable subproblems, enabling process-level supervision within a
      single on-policy rollout. This keeps the model exploring near the boundary of
      its current capability, making hard problems more effective for training.
    \item \textbf{Subproblem-level normalization for fine-grained credit assignment.}
      We introduce \emph{subproblem-level normalization}, which normalizes rewards
      independently at each subproblem position and assigns the resulting advantages to
      the corresponding answer spans, enabling fine-grained credit assignment without
      external rubrics or additional reward models.
    \item \textbf{Theoretical and empirical validation.}
      We provide a metric recovery analysis showing that subproblem decomposition
      lifts hard problems out of gradient dead zones, with larger relative gains as
      the original problem becomes harder. Experiments across seven mathematical
  reasoning benchmarks verify these predictions and show consistent gains over
  strong baselines (+4.1/+1.9 average-point gains on Qwen3-4B/14B; +3.7 pass@$1$
  and +4.6 pass@$64$ points on three hard benchmarks).
  \end{itemize}
\section{Related Work}
\label{sec:related}

\paragraph{Reinforcement Learning with Verifiable Rewards (RLVR)}

Recent advances in Large Language Models (LLMs) have highlighted the effectiveness of Reinforcement Learning (RL) in domains with deterministic verifiers such as mathematics and programming \cite{DeepSeekMath, jaech2024openai,trinh2024solving,yang2024qwen2,Qu2025RLADTL,wang2025beyond}. Unlike open-ended generation, these tasks provide unambiguous feedback, allowing for the optimization of policy models through algorithms like Proximal Policy Optimization (PPO) \cite{schulman2017proximal} or the more memory-efficient Group Relative Policy Optimization (GRPO) \cite{deepseekr1}. However, RLVR faces a significant challenge: for difficult problems, the reward signal becomes extremely sparse, leading to a failure in obtaining meaningful policy gradients \cite{uesato2022solving}. This challenge is often framed as a credit assignment problem: outcome-based rewards provide a global signal but fail to pinpoint which specific reasoning steps contributed to the final success or failure \cite{lightman2023let}. While iterative self-improvement methods like STaR \cite{zelikman2022star} and ReST \cite{gulcehre2023reinforced,zhang2024rest} attempt to bridge this gap through rejection sampling on easier instances, they still struggle when the task's difficulty exceeds the model's current exploration horizon. Consequently, curriculum learning \cite{bengio2009curriculum,Yang2025FromIT,Li2025AdaCuRLAC,Parashar2025CurriculumRL,Wu2025ProgressiveMC} has become a common way to densify learning signals for hard problems by breaking hard tasks into manageable stages.
\paragraph{Curriculum Learning for Reasoning}
Existing curriculum learning methods for mathematical reasoning can be broadly categorized into two paradigms. The first category focuses on providing external hints or guidance when the model fails to solve a challenging problem. Notable works such as StepHint \cite{zhang2025stephint}, Scaf-GRPO \cite{zhang2025scaf}and other hint-driven RL frameworks \cite{wu2025thought,qiyuan2026hipo,qu2026pope,Luffy,Shi2026R3LRR}, utilize teacher model or self-generated rationales as auxiliary prefixes to lower the exploration threshold. The second category involves rewriting the original problem into simpler versions or augmenting the prompt with supplementary information to facilitate reasoning \cite{chen2026cog,Wu2025ProgressiveMC,li2026mathmixup,dai2026harder,li2026questa,Liang2025BeyondPS}. As seen in MQR \cite{dai2026harder} and QuestA \cite{li2026questa}, these methods effectively create a difficulty gradient by manipulating the problem context. However, a fundamental limitation shared by these methods is their reliance on additional context. By providing the hint or reformulated problem as a static prefix, these approaches primarily optimize the model's continuation capability. As a result, the model fails to internalize the underlying scaffolding logic, as it is never required to generate the hints or auxiliary structures itself. In contrast, SCRL requires the model to generate the entire scaffolded multi-part sequence within a structured response, ensuring that the policy learns to both construct the intermediate reasoning steps and solve the final target problem.
\section{Method}
\label{sec:method}
We propose \textbf{SCRL} (\textbf{Subproblem Curriculum Reinforcement Learning}),
  a curriculum RL framework that turns hard problems into verifiable subproblem
  curricula for finer-grained credit assignment. SCRL has three steps. First, given
  a reference solution, an external LLM derives $K$ verifiable subproblems from
  the reasoning chain and constructs the subproblem curriculum. Second, the policy answers all $K$ subproblems in one on-policy rollout. We then
  verify each subproblem answer and apply progress-aware correction to obtain
  progress-aware subproblem rewards. Subproblem-level normalization computes an
  advantage for each subproblem position, which is then used for token-level credit
  assignment. Finally, to reduce prompt mismatch,
  SCRL uses mixed-group training, jointly optimizing curriculum rollouts and
  original-problem rollouts in the same update.
  \subsection{Preliminaries: GRPO}
  \label{sec:prelim}

  Given a prompt $q$, GRPO samples $G$ rollouts
  $\{o_i\}_{i=1}^{G}\sim\pi_\theta(\cdot\mid q)$ and assigns each rollout a scalar
  verifiable reward $r_i$. It then optimizes the clipped objective
  \begin{equation}
  \mathcal{L}_{\mathrm{GRPO}}(\theta) =
  -\frac{1}{\sum_i L_i}
  \sum_{i=1}^{G} \sum_{t=1}^{L_i}
  \min\!\Bigl(
  \rho_{i,t}A_i,\;
  \mathrm{clip}(\rho_{i,t},1-\varepsilon,1+\varepsilon)A_i
  \Bigr)
  - \beta D_{\mathrm{KL}}(\pi_\theta\|\pi_{\mathrm{ref}}).
  \label{eq:loss_grpo}
  \end{equation}

Here $A_i=\frac{r_i-\mathrm{mean}(\{r_i\}_{i=1}^{G})}
  {\mathrm{std}(\{r_i\}_{i=1}^{G})}$ is the group-normalized advantage, and 
  $\rho_{i,t}=\frac{\pi_\theta(o_{i,t}\mid q,o_{i,<t})}
  {\pi_{\theta_{\mathrm{old}}}(o_{i,t}\mid q,o_{i,<t})}$ is the importance sampling ratio at token $t$.
  Since the same $A_i$ is assigned to every token in $o_i$, GRPO performs
  sample-level credit assignment.
\subsection{SCRL Framework}
\label{sec:framework}

\paragraph{Build subproblems.}
For each hard problem $x$, we start from an existing chain-of-thought reference
  solution. An external LLM rewrites its intermediate progress nodes into $K$
  verifiable subproblems, rather than solving the problem from scratch. The exact
  generation prompt is provided in Appendix~\ref{app:prompt_for_subproblems}, and
  the main guidelines are summarized below.

  \begin{tcolorbox}[
      colback=cyan!2!white,
      colframe=cyan!80!black,
      title=Guidelines for Subproblem Generation,
      fonttitle=\bfseries,
      arc=1.5mm,
      breakable,
      enhanced,
      left=2mm, right=3mm, top=2mm, bottom=2mm,
      boxrule=0.6pt
  ]
  \small

  \noindent\textbf{1. Increasing difficulty.}
  Subproblems are ordered from easier to harder, with the final subproblem fixed as
  the original problem:
  \[
  s^{(1)} \prec s^{(2)} \prec \cdots \prec s^{(K)} = x .
  \]

  \noindent\textbf{2. Linked but self-contained.}
  Earlier subproblems provide useful intermediate results for later ones, while
  each subproblem remains a complete standalone question.

  \vspace{0.4em}
  \noindent\textbf{3. Verifiable answers.}
    Each subproblem has an objectively checkable answer for independent verification.

  \end{tcolorbox}
\paragraph{Curriculum prompt.}
Let $x$ denote the original problem. We define the curriculum prompt
  $t_K(x)$ as the prompt that presents all $K$ subproblems
  $s^{(1)},\ldots,s^{(K)}$ simultaneously and asks the model to solve them in order. Thus, $x$
  corresponds to the original-problem rollout, while $t_K(x)$ corresponds to the
  curriculum rollout. The detailed prompt template is provided in Appendix~\ref{app:curriculum_prompt}.
\vspace{-2pt}
\paragraph{Response format.}
During curriculum rollouts, the model is asked to answer the $K$ subproblems
  using explicit tags $\texttt{<pj>}$ and $\texttt{</pj>}$:
\[
\texttt{<p1>}\;a^{(1)}\;\texttt{</p1>}
\;\cdots\;
\texttt{<pK>}\;a^{(K)}\;\texttt{</pK>},
\]
where $a^{(j)}$ is the response to subproblem $j$.
These tags not only specify the response format, but also mark the token span of
  each subproblem answer.
  This allows us to verify each answer separately and later assign the
  corresponding subproblem-level advantage back to the tokens inside that span.

\begin{figure*}[t]
      \centering
      \includegraphics[width=\textwidth]{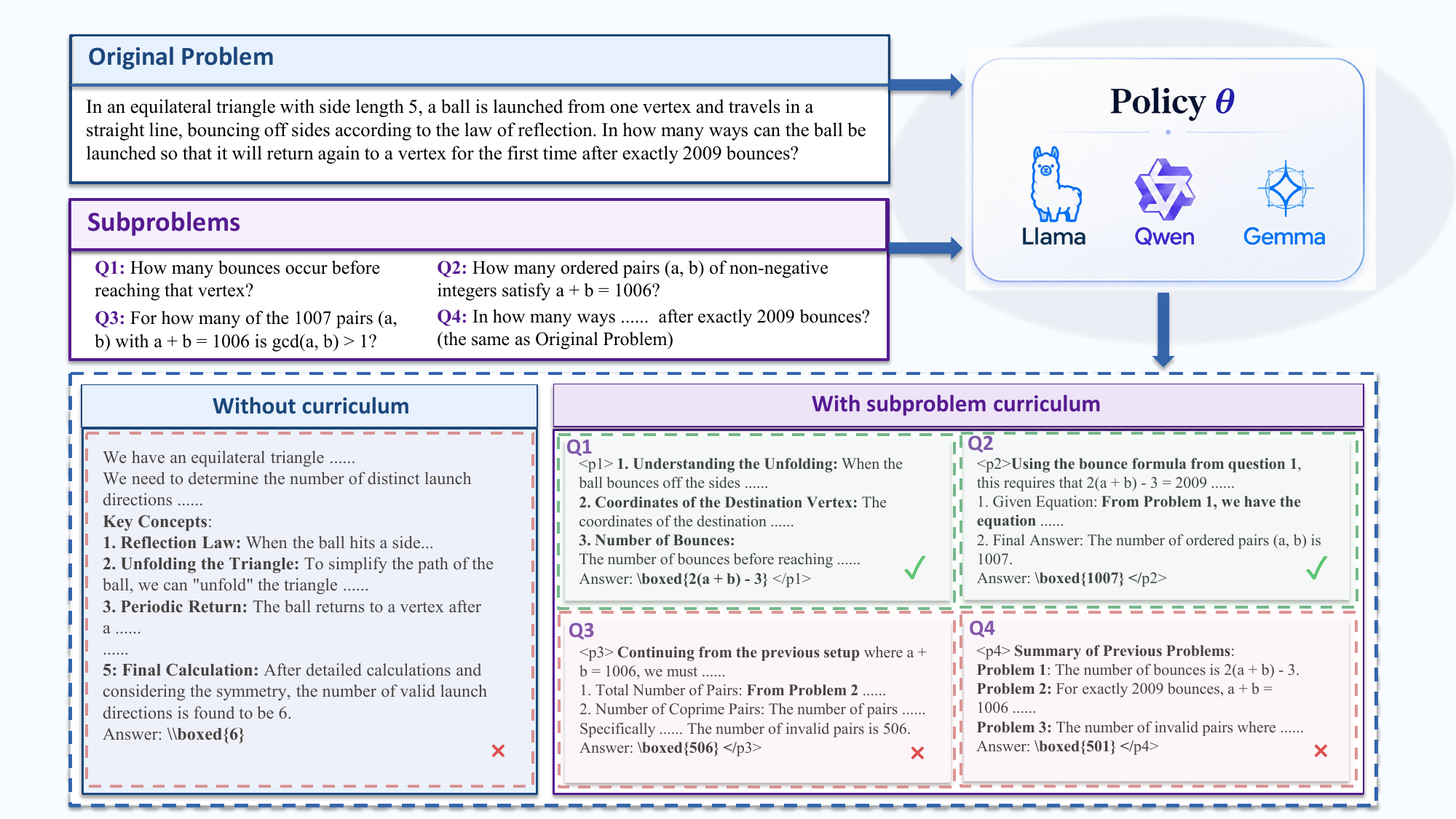}

 \caption{
  Illustration of mixed training rollouts. The policy generates both
  original-problem rollouts and curriculum rollouts. The tagged response format
  \texttt{<pj>}...\texttt{</pj>} in curriculum rollouts identifies the answer span of each subproblem,
  enabling finer-grained credit assignment.
  }
      \label{fig:scrl_rollout_example}
  \end{figure*}

\subsubsection{Progress-Aware Subproblem Rewards}
\label{sec:subproblem_grpo}
\label{sec:process_correct}
\paragraph{Curriculum progress.}
  For a curriculum rollout $o_i \sim \pi_\theta(\cdot \mid t_K(x))$, verifying the $K$ extracted subproblem answers gives a raw
  reward vector
  $\mathbf{r}_i=(r_i^{(1)},\ldots,r_i^{(K)})\in\{0,1\}^K$.
  If the response does not follow the required format, we set
  $\mathbf{r}_i=\mathbf{0}$.
  We define the curriculum progress $k_i \in \{0,1,\ldots,K\}$ as the maximum number of
  consecutively solved subproblems from the beginning:
  \begin{equation}
  k_i :=
  \max\bigl\{j \in \{0,1,\ldots,K\}
  \;\big|\;
  r_i^{(1)} = \cdots = r_i^{(j)} = 1
  \bigr\}.
  \label{eq:progress}
  \end{equation}
  Thus, $k_i=0$ means the first subproblem is incorrect, while $k_i=K$ means all
  subproblems are solved. The curriculum progress $k_i$ tracks the current policy's capability boundary on
    the hard problem, and also identifies the intermediate progress actually achieved by the rollout.
    \vspace{-2pt}
\paragraph{Progress-aware correction.}
  Directly rewarding each subproblem independently may credit later subproblems despite earlier
  failures, creating a potential reward-hacking shortcut. We therefore align
  rewards with curriculum progress by keeping only the consecutively solved prefix:
  \begin{equation}
  \tilde{r}_i^{(j)} :=
  \begin{cases}
  r_i^{(j)}, & j \le k_i, \\
  0,          & j >  k_i,
  \end{cases}
  \qquad
  \tilde{\mathbf{r}}_i =
  \bigl(\tilde{r}_i^{(1)}, \tilde{r}_i^{(2)}, \ldots,\tilde{r}_i^{(K)}\bigr).
  \label{eq:pca}
  \end{equation}
  For example, when $K=4$, $(1,1,0,1)$ is corrected to $(1,1,0,0)$.
  For notational convenience, we use $R_i^{(j)} := \tilde{r}_i^{(j)}$ as the final
  subproblem reward for training.

\subsubsection{SCRL Training Algorithm}
  \label{sec:col_grpo}
  \label{sec:v8}

  In this section, we describe the training details of SCRL, including
  subproblem-level normalization for advantage computation, token-level credit
  assignment, and mixed-group training. The full training procedure is summarized
  in Appendix~\ref{app:scrl_algorithm}.
\paragraph{Subproblem-level normalization.}
Given $G$ curriculum rollouts $o_i$ for
  $i=1,\ldots,G$, we normalize the final subproblem rewards at each subproblem
  position $j$ across the rollout group:
\begin{equation}
  A_i^{(j)} =
  \frac{
  R_i^{(j)} - \mathrm{mean}\bigl(\{R_i^{(j)}\}_{i=1}^{G}\bigr)
  }{
  \mathrm{std}\bigl(\{R_i^{(j)}\}_{i=1}^{G}\bigr)
  }.
  \label{eq:col_grpo}
  \end{equation}
Thus, the subproblem-level advantage $A_i^{(j)}$ measures the relative success of rollout $i$ at
  subproblem position $j$ within the rollout group,
  independent of rewards at other subproblem positions.
\vspace{-2pt}
\paragraph{Token-level credit assignment.}
After computing the subproblem-level advantages, we assign them back to the
  tokens of the corresponding subproblem answers. Using the structured response
  format, we define $\mathrm{sub}_i(t)=j$ if token $o_{i,t}$ lies between
  \texttt{<pj>} and \texttt{</pj>}; then
  $A_{i,t}=A_i^{(\mathrm{sub}_i(t))}$ gives the token-level advantage. Tokens outside all answer spans receive zero
  advantage, and if the response does not follow the required format, all tokens in
  that response receive zero advantage. This converts subproblem-level progress
  into token-level learning signals for the corresponding answer spans.
\vspace{-2pt}
\paragraph{Mixed-group training.}
  \label{sec:training}
  Training only on the curriculum prompt $t_K(x)$ can cause prompt mismatch,
  because evaluation uses the original prompt $x$. We therefore use mixed-group
  training: for each problem $x$, $G/2$ rollouts are sampled from $t_K(x)$ and
  optimized with token-level advantages from subproblem-level normalization, while
  the other $G/2$ rollouts are sampled from $x$ and optimized with standard
  outcome-based GRPO. The final SCRL objective is
 \begin{align}
  \mathcal{L}_{\mathrm{SCRL}}(\theta)
  &= -\frac{1}{\sum_{i=1}^{G} L_i}
  \Bigg[
  \sum_{i=1}^{G/2} \sum_{t=1}^{L_i}
  \min\!\Bigl(\rho_{i,t}A_{i,t},
  \mathrm{clip}(\rho_{i,t},1-\varepsilon,1+\varepsilon)A_{i,t}\Bigr)
  \notag \\
  &\quad
  +\sum_{i=G/2+1}^{G} \sum_{t=1}^{L_i}
  \min\!\Bigl(\rho_{i,t}A_i,
  \mathrm{clip}(\rho_{i,t},1-\varepsilon,1+\varepsilon)A_i\Bigr)
  \Bigg]
  -\beta D_{\mathrm{KL}}(\pi_\theta \,\|\, \pi_{\mathrm{ref}}).
  \label{eq:loss_scrl_full}
  \end{align}
  The two bracketed terms correspond to curriculum rollouts and original-problem
  rollouts respectively. The complete training procedure is summarized in Algorithm~\ref{alg:scrl_training}.

  \section{Theoretical Analysis}
  \label{sec:theory}

Using the information geometry of the policy manifold
  $\mathcal{M}=\{\pi_\theta:\theta\in\Theta\}$ equipped with the Fisher--Rao
  metric~\citep{amari2016information}, we show that hard problems can place
  outcome-based GRPO in a \emph{gradient dead zone}, while subproblem
  decomposition lifts optimization to a product manifold that recovers useful
  gradient information. Full discussions and proofs are provided in
  Appendix~\ref{app:proofs}.

\begin{definition}[Effective and Lifted Gradient Information Matrices]
  \label{def:egim}
  Under GRPO, let $o_1,\ldots,o_G \overset{\mathrm{i.i.d.}}{\sim} \pi_\theta(\cdot\mid x)$
be $G$ sampled rollouts. The
  \textbf{effective gradient information matrix} (EGIM) of $x$ and the
  \textbf{lifted EGIM} of its subproblem transformation $\mathcal{T}(x)=t_K(x)$ are
  \begin{align}
  \bm{F}_x(\theta)
  &=
  \frac{1}{G}\sum_{i=1}^{G}
  \mathbb{E}\!\left[g_i(x)g_i(x)^\top\right],
  \qquad
   \text{where } g_i(x)=\hat{A}_i(x)\nabla_\theta\log\pi_\theta(o_i\mid x),
  \label{eq:egim}
  \\
  \bm{F}_{\mathcal{T}(x)}(\theta)
  &=
  \frac{1}{K}\sum_{j=1}^{K}
  \frac{1}{G}\sum_{i=1}^{G}
  \mathbb{E}\!\left[
  \hat{A}_i^{(j)2}
  \nabla_\theta\log\pi_\theta(o_i^{(j)}\mid t_K(x))
  \nabla_\theta\log\pi_\theta(o_i^{(j)}\mid t_K(x))^\top
  \right],
  \label{eq:lifted_egim}
  \end{align}
Here $\hat{A}_i(x)$ and $\hat{A}_i^{(j)}$ denote the original-problem and
  subproblem-position advantages respectively, and the smallest eigenvalue $\lambda_{\min}(\bm{F}_{\mathcal{T}(x)}(\theta))$ measures
  the weakest useful gradient signal.
  \end{definition}

  \begin{theorem}[Gradient Dead Zone]
  \label{thm:dead_zone}
  Let $p(x;\theta)=\Pr_{\pi_\theta}[r(x,o)=1]$ be the probability that the current
  policy solves $x$. If $p(x;\theta)<\delta$, then
  \begin{equation}
      \lambda_{\min}\!\left(\bm{F}_x(\theta)\right)
      \leq
      G\delta\cdot C_{\hat{A}}^2\cdot B_s^2
      =
      O(\delta),
      \label{eq:dead_zone_bound}
  \end{equation}
  \end{theorem}

where $C_{\hat{A}}\leq \sqrt{G-1}$ bounds the normalized advantage magnitude and $B_s$ bounds the score norm (both derived in Appendix~\ref{app:adv_bound}).

Theorem~\ref{thm:dead_zone} shows that direct RLVR training becomes
  ineffective on hard problems: when correct rollouts are rare, reward groups
  collapse and the worst-case effective gradient signal vanishes.

\begin{theorem}[Metric Recovery via Subproblem Decomposition]
  \label{thm:recovery}
  Let $x$ be in the gradient dead zone with $p(x;\theta)<\delta$. Suppose the
  subproblem construction satisfies
  \[
  p_j(x;\theta):=\Pr[R^{(j)}=1\mid t_K(x)]\in[p^\star,1-p^\star]
  \quad \text{for all } j<K,
  \]
  where $p^\star\in(\delta,1/2]$. Under the conditional identifiability assumption
($\mathbb{E}[(v^\top\nabla_\theta\log\pi_\theta)^2 \mid r{=}r_0]
\geq \sigma_{\min}^2 > 0$ for all unit $v$, $r_0\in\{0,1\}$),
  \begin{align}
  \lambda_{\min}\!\left(\bm{F}_{\mathcal{T}(x)}(\theta)\right)
  &\geq
  \frac{1}{K}\,c(p^\star,G,\sigma_{\min}) > 0,
  &
  \frac{
  \lambda_{\min}(\bm{F}_{\mathcal{T}(x)}(\theta))
  }{
  \lambda_{\min}(\bm{F}_x(\theta))
  }
  &= \Omega\!\left(\frac{1}{\delta}\right),
  \end{align}
  where $c(p^\star,G,\sigma_{\min})
  =
  \bigl(1-(p^\star)^G-(1-p^\star)^G\bigr)\sigma_{\min}^2$
  is a positive constant independent of $\delta$.
  \end{theorem}

  Theorem~\ref{thm:recovery} shows that subproblem curriculum helps
  hard problems by recovering a non-degenerate learning geometry, even when the
  original problem provides almost no useful gradient signal. Moreover, the
  recovery ratio grows as $p(x;\theta)\to 0$, predicting larger relative gains on
  harder problems.
\section{Experiment}
\subsection{Experimental Setup}
\paragraph{Models.}
To investigate the scalability and effectiveness of our proposed method across different model capacities, we conduct experiments on the Qwen and Llama series. Specifically, we utilize \texttt{Qwen3-4B-Base}, \texttt{Qwen3-14B-Base} and \texttt{Llama3.2-3B-Instruct} as our base policies. 
\vspace{-5pt}
\paragraph{Training Setup.}
We use the training set \texttt{hard\_1024}, a subset of 1,024 problems randomly selected from the high-difficulty competition mathematics dataset provided by \cite{yang2026int}. For SCRL, subproblems are generated with the DeepSeek-V3.2
  API with $K=4$. All models are trained using the \textbf{Verl} framework~\cite{verl} for a total of 300 steps. Detailed hyperparameter configurations are provided in Appendix~\ref{app:hyperparameters}.
  \vspace{-5pt}
\paragraph{Benchmark.}
We evaluate the models on seven widely used mathematical reasoning benchmarks: \texttt{OlympiadBench}, \texttt{Minerva}, \texttt{MATH-500}, \texttt{AIME 2024}, \texttt{AIME 2025}, \texttt{AMC}, and \texttt{IMO-Bench}.
\vspace{-5pt}
\paragraph{Baseline Settings.} 
We compare our method against the following competitive baselines: \texttt{SFT}, \texttt{GRPO}~\cite{deepseekr1},\texttt{DAPO}~\cite{Yu2025DAPOAO},\texttt{QuestA}~\cite{li2026questa} and \texttt{NuRL}~\cite{chen2025nudging}.Implementation details are provided in Appendix~\ref{app:baseline_settings}. 
\subsection{Main Results and Further Analysis}

\begin{table*}[h]
\caption{\textbf{Main Results} on mathematical reasoning benchmarks. The best results are highlighted in \textbf{bold}, and the second-best
  results are \underline{underlined}.}
  \label{tab:main_results}
  \centering
  \small
  \setlength{\tabcolsep}{6pt} 
  \resizebox{\textwidth}{!}{
  \begin{tabular}{l|l|c|c|c|c|c|c|c|c}
    \toprule
    \textbf{Base Model} & \textbf{Method} & \textbf{Olym.B} & \textbf{Minerva} & \textbf{MATH} & \textbf{AIME'24} & \textbf{AIME'25} & \textbf{AMC} & \textbf{IMO-B} & \textbf{AVG}\\
    \midrule

    \multirow{7}{*}{Llama3.2-3B-Instruct} 
    & Initial Model & 13.1 & 13.7 & 44.0 & 6.4 & \underline{0.7} & 20.6 & 4.5 & 14.7 \\
    & SFT           & 13.1 & 11.0 & 44.1 & 2.5 & \underline{0.7}& 18.1 & 2.6 &  13.2\\
    & GRPO          & 13.7 & 14.9 & 44.5 & \textbf{10.3} & 0.5 & 20.9 & 5.4 & 15.7 \\
    & DAPO          & 14.6 & \underline{15.1}& \textbf{45.9 }& 9.8 & 0.3 & 22.2 & 5.8 & \underline{16.2}\\
    & QuestA        & 14.6 & 14.8 &\textbf{ 45.9 }& 8.5 & 0.4 & \underline{21.4} & \underline{6.1}&  16.0\\
    & NuRL          &\underline{15.1} & 14.9 & \underline{45.2} & \underline{10.2}& 0.3 & \textbf{21.7} & 5.9 & \underline{16.2}\\
    & \textbf{SCRL} & \textbf{15.2} & \textbf{15.2} & \underline{45.2} & \textbf{10.3} & \textbf{0.8}& \underline{21.4} & \textbf{6.8} &\textbf{16.4}  \\
    \midrule

    \multirow{7}{*}{Qwen3-4B-Base} 
    & Initial Model & 29.0 & 15.6 & 64.4 & 10.9 & 6.4 & 38.1 & 6.1 &  24.4\\
    & SFT           & 31.1 & 25.4 & 71.0 & 10.6 & 10.0 & 41.0 & 7.1 & 28.0\\
    & GRPO          & 35.1 & 29.7 & 76.2 & \underline{14.5}& 7.2 & 46.1 & 7.8 & 30.9\\
    & DAPO          & 34.9 & 30.1 & 76.0 & 11.8 & 9.3 & \underline{46.8}& \underline{8.3}& 31.0 \\
    & QuestA        & 35.0 & \underline{32.0}& 75.9 & 14.1 & \underline{11.7}& \underline{46.8} & \underline{8.3}& \underline{32.0}\\
    & NuRL          & \underline{35.6}& 30.3 & \underline{77.6}& 13.2 & 8.4 & 45.4 & 7.1 & 31.1\\
    & \textbf{SCRL} & \textbf{39.2} & \textbf{33.1} & \textbf{79.8} & \textbf{16.5} & \textbf{15.3} & \textbf{52.4} & \textbf{8.7} & \textbf{35.0} \\
    \midrule

    \multirow{7}{*}{Qwen3-14B-Base} 
    & Initial Model & 35.3 & 23.7 & 75.7 & 14.4 & 10.7 & 50.2 & 7.4  & 31.1  \\
    & SFT           & \textbf{41.2} & 24.3 & 72.0 & 20.0 &\textbf{ 20.0 }& 49.3 & \underline{11.0}& 34.0\\
    & GRPO          & 38.4 & 33.4 & 81.3 & 21.6 & 11.9 & 58.1 & 9.9  & 36.4 \\
    & DAPO          & 37.5 & 32.6 & 79.9 & 22.7 & 12.3 & 59.2 & 9.9 & 36.3 \\
    & QuestA        & 40.4 & \underline{34.0}& \underline{82.6}& 18.5 & 11.6 & \textbf{60.2 }& 10.4  & \underline{36.8}\\
    & NuRL          & 38.3 & 32.6 & 82.0 & \underline{23.6} & 11.2 & 58.1 & 9.4  & 36.5 \\
    & \textbf{SCRL} & \underline{40.9}& \textbf{34.7} & \textbf{83.2} & \textbf{24.4} & \underline{13.1} & \underline{60.0} & \textbf{11.9}  &  \textbf{38.3}\\
    \bottomrule
  \end{tabular}
  }
\end{table*}

\begin{figure}[htbp]
  \centering
  \begin{minipage}{0.32\textwidth}
      \centering
      \includegraphics[width=\linewidth]{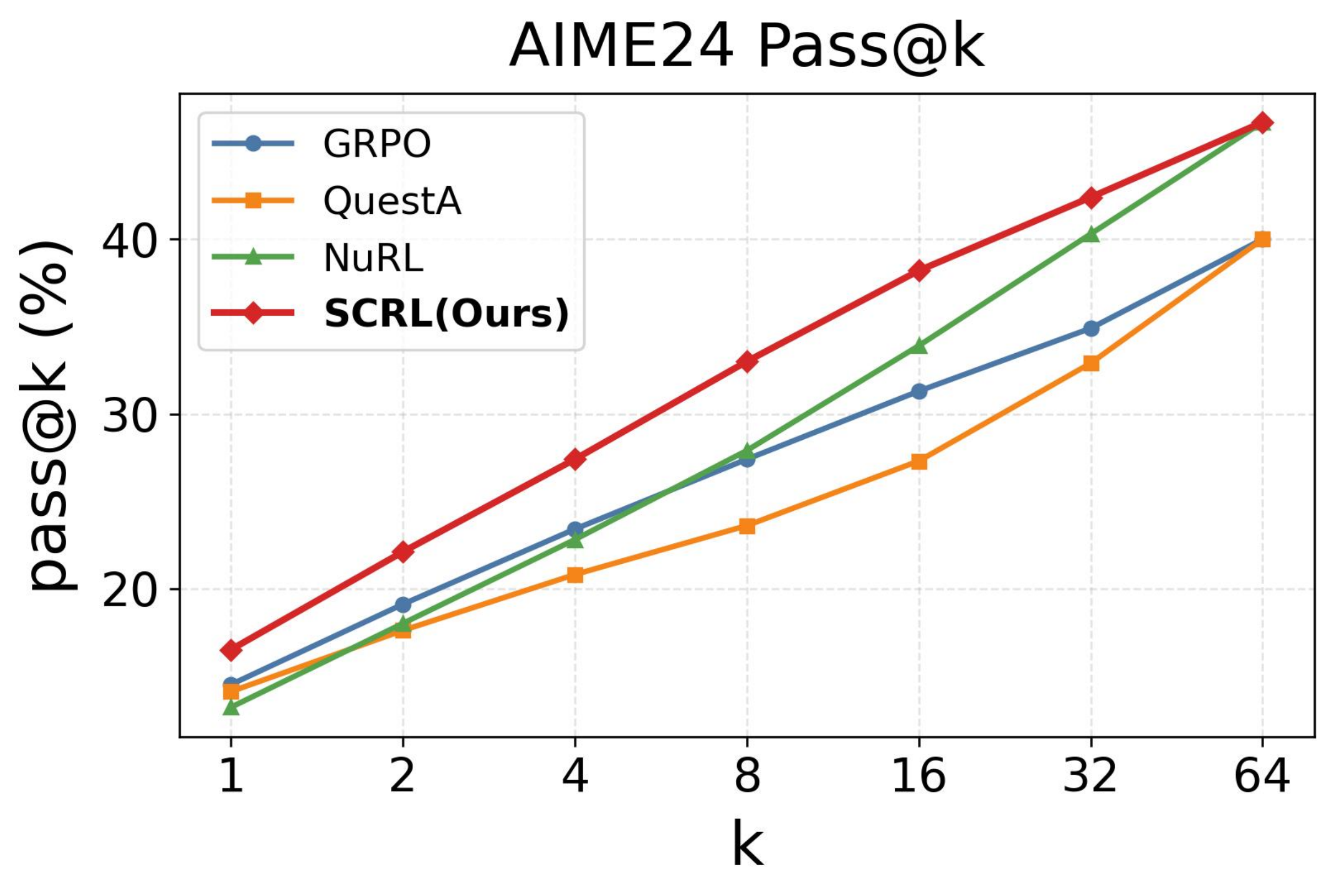}
  \end{minipage}
  \hfill
  \begin{minipage}{0.32\textwidth}
      \centering
      \includegraphics[width=\linewidth]{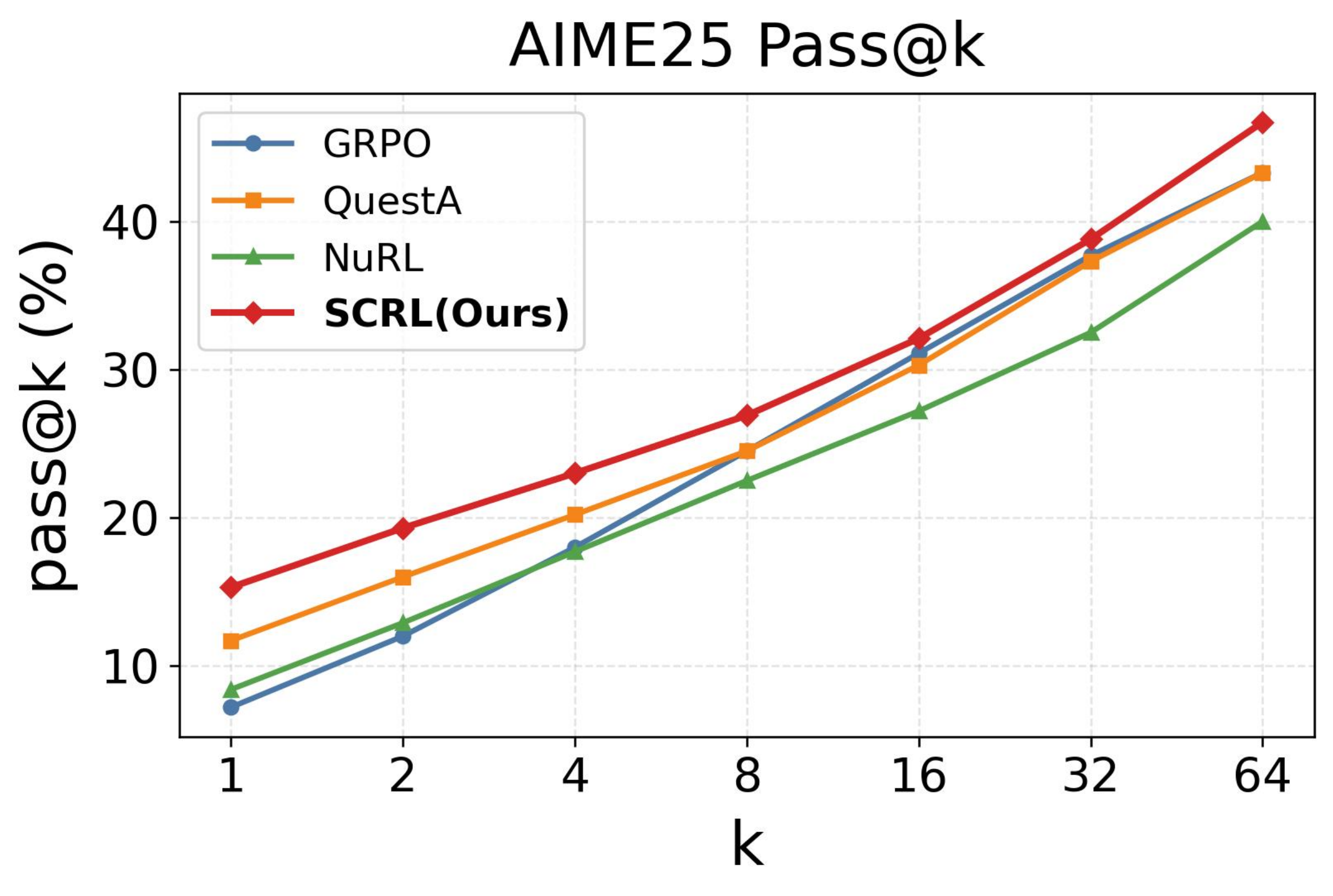}
  \end{minipage}
  \hfill
  \begin{minipage}{0.32\textwidth}
      \centering
      \includegraphics[width=\linewidth]{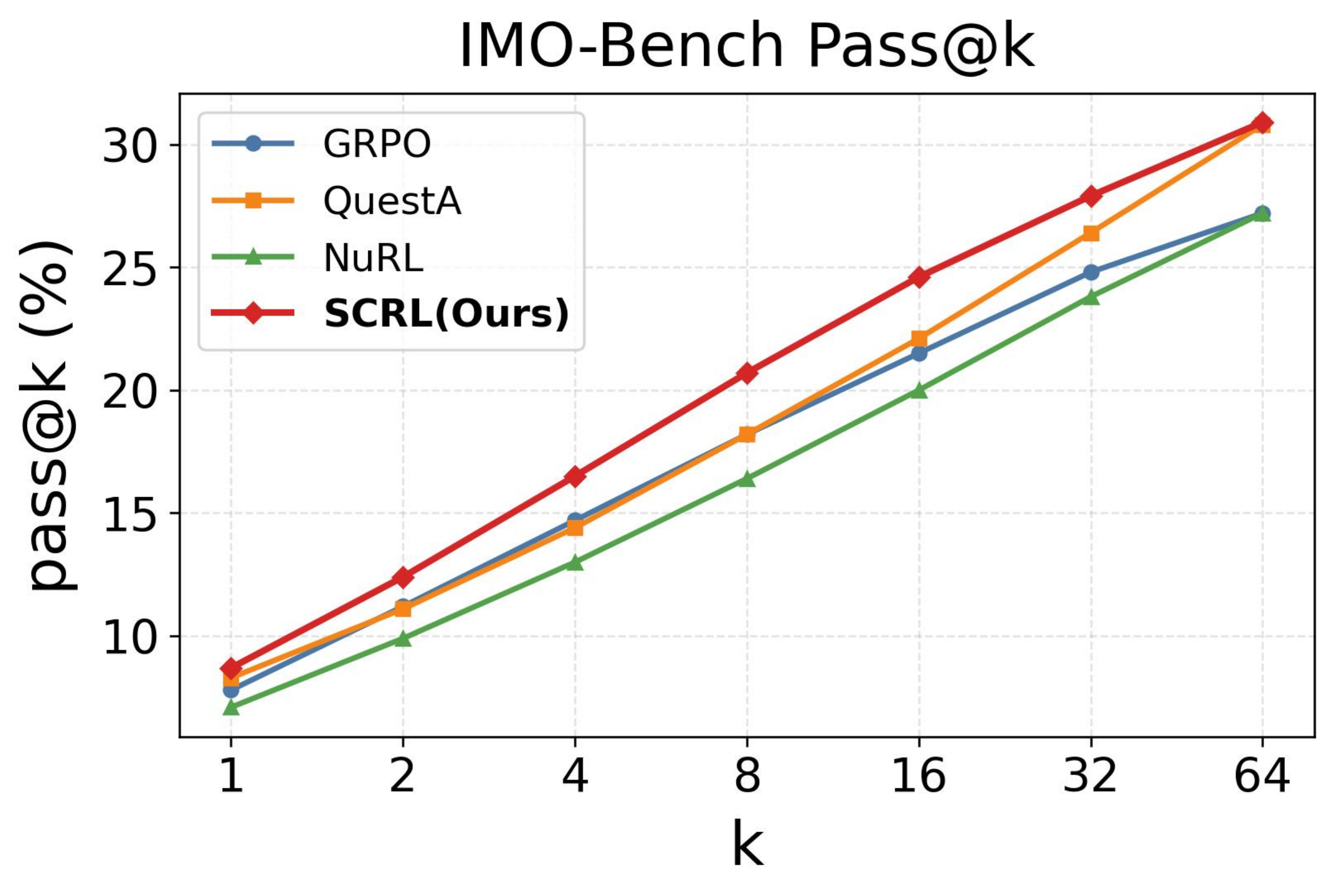}
  \end{minipage}
  \caption{Pass@$k$ curves on AIME24, AIME25, and IMO-Bench on Qwen3-4B-Base.}
  \label{fig:passk_hard_benchmarks}
  \end{figure}

The main results across seven mathematical reasoning benchmarks are summarized
  in Table~\ref{tab:main_results}, with full experimental results provided in
  Appendix~\ref{app:full_results}. 
  \paragraph{Superior Performance Across All Benchmarks.}
  As shown in Table~\ref{tab:main_results}, \textbf{SCRL} consistently outperforms
  vanilla GRPO and competitive baselines including DAPO, QuestA, and NuRL across
  three model scales: Llama3.2-3B, Qwen3-4B, and Qwen3-14B. In terms of average
  accuracy (Avg), SCRL achieves the best performance in all settings. The
  gain is especially clear on Qwen3-4B, where SCRL reaches an average score of
  \textbf{35.0\%}, improving over the second-best baseline QuestA
  (\textbf{32.0\%}) by \textbf{3.0} points and over vanilla GRPO
  (\textbf{30.9\%}) by \textbf{4.1} points. On challenging benchmarks such as
  AIME'25, SCRL also shows strong gains, achieving \textbf{15.3\%} compared with
  QuestA's \textbf{11.7\%}.
\vspace{-5pt}
  \paragraph{Curriculum progress transfers to hard-problem solving.}
  Figure~\ref{fig:passk_hard_benchmarks} shows pass@$k$ curves on AIME24, AIME25,
  and IMO-Bench. SCRL consistently outperforms GRPO and other curriculum RL baselines across the
  entire evaluated range of $k$, indicating stronger hard-problem solving ability.

\begin{figure}[htbp]
\centering
\begin{minipage}[t]{0.48\textwidth}
      \centering
      \includegraphics[width=\linewidth]{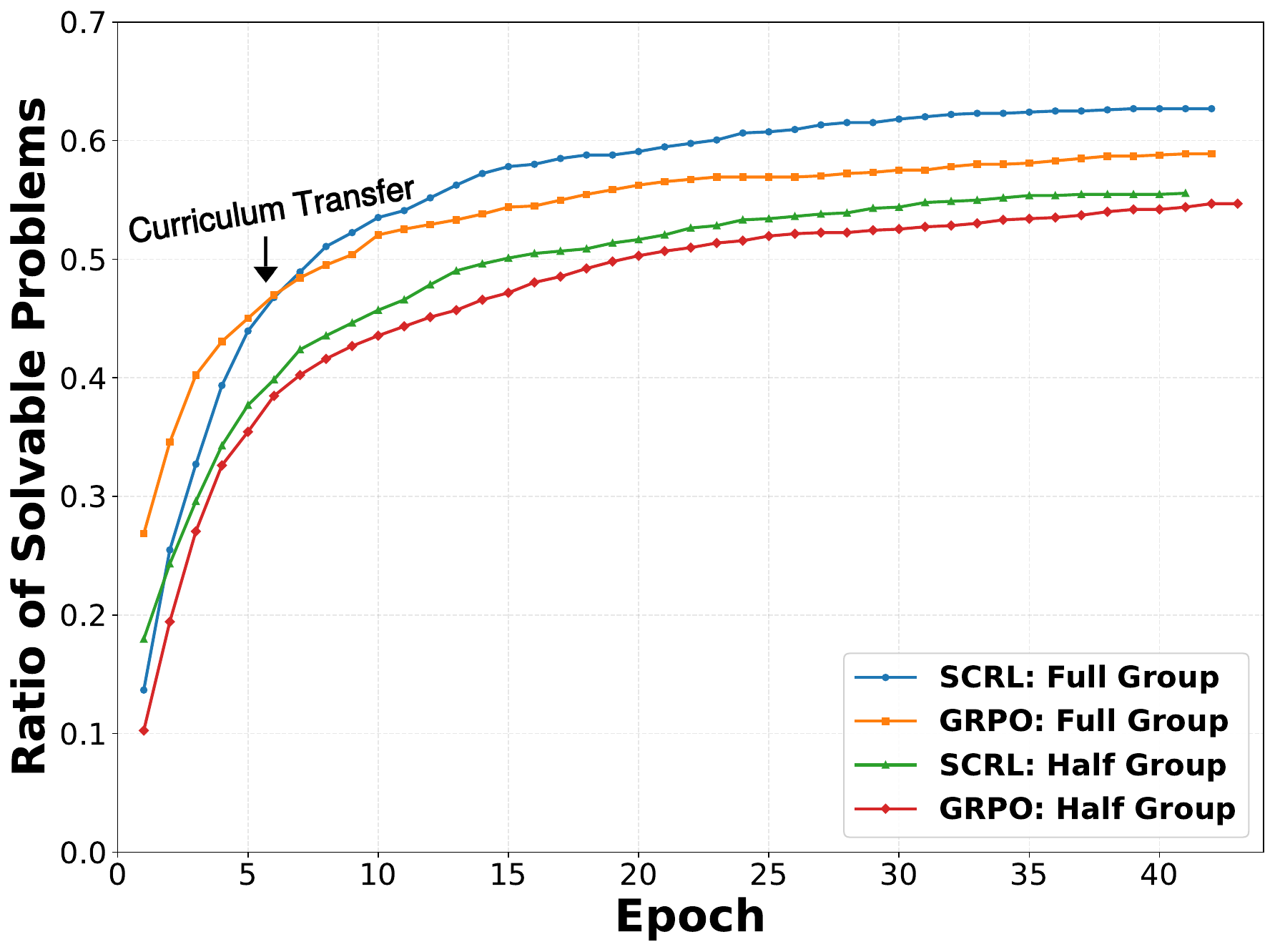}
       \setlength{\abovecaptionskip}{-10pt} 
      \caption{
      Ratio of solvable problems during training of Qwen3-4B-Base.
      }
      \label{fig:solvable_set_ratio}
\end{minipage}
\hfill
\begin{minipage}[t]{0.48\textwidth}
      \centering
      \includegraphics[width=\linewidth]{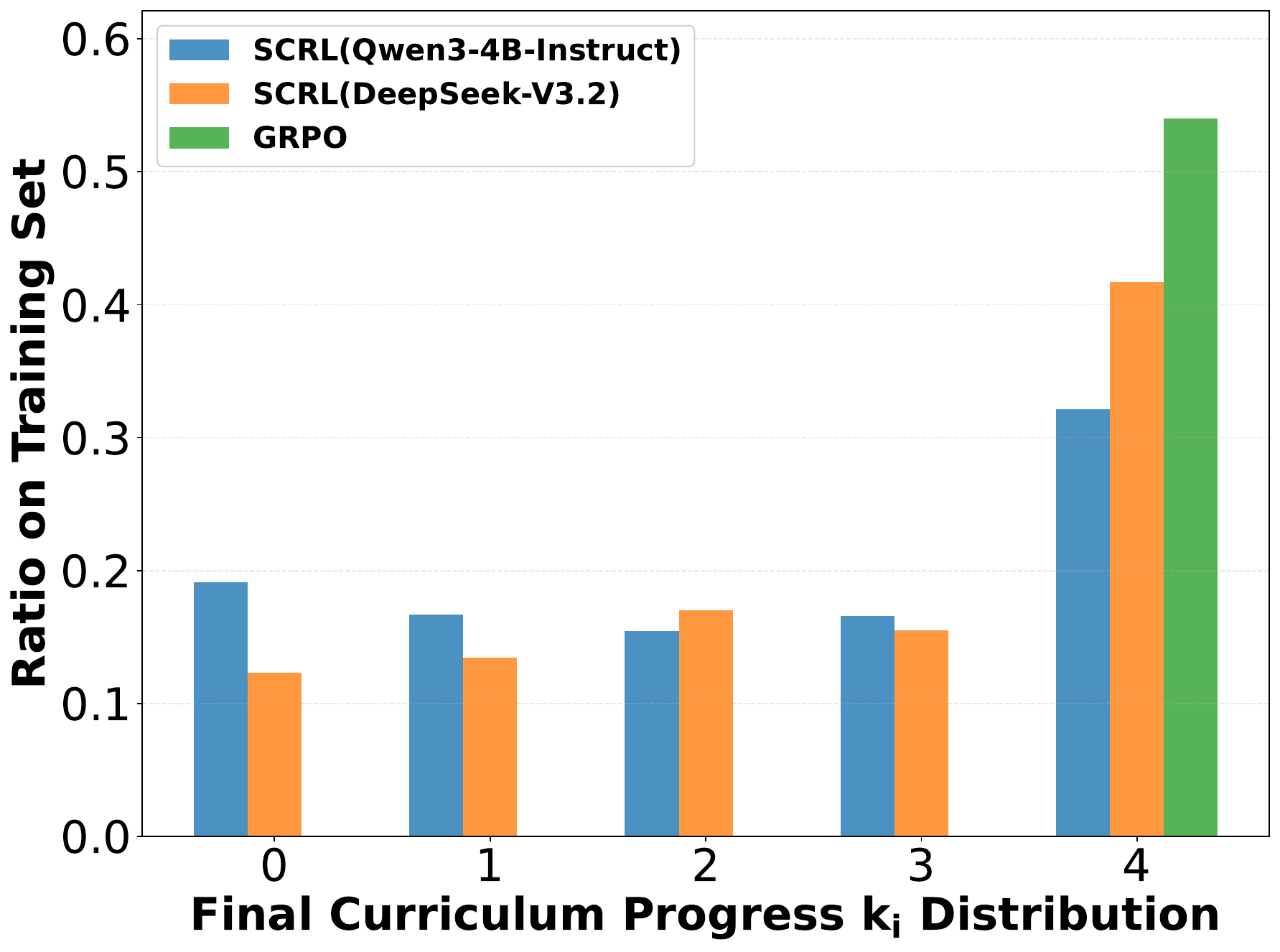}
       \setlength{\abovecaptionskip}{-10pt} 
      \caption{
      Final curriculum-progress distribution on the training set. 
      }
      \label{fig:curriculum_progress_ratio_with_grpo}
\end{minipage}
\end{figure}

  Figure~\ref{fig:solvable_set_ratio} further tracks the ratio of solvable
  problems during training, where a problem is counted as solvable once it is fully
  solved at least once. The \emph{full group} statistic counts success in either
  the original-problem or curriculum format, while the \emph{half group} statistic
  uses only half-budget original-problem rollouts, matching SCRL's mixed-group
  setting. SCRL achieves a higher solvable ratio than GRPO under both protocols,
  showing that curriculum progress transfers back to direct hard-problem solving
  rather than only improving curriculum-format rollouts.
\vspace{-5pt}
 \paragraph{\textbf{SCRL does not rely on highly curated subproblems.}}
  We further examine whether SCRL depends on high-quality subproblem construction.
  Table~\ref{tab:ablation_generator_main} compares subproblems generated by
  DeepSeek-V3.2 and a weaker Qwen3-4B-Instruct generator, using the same generation
  prompt and downstream training pipeline. In both cases, the generator is given
  the dataset reference solution, so it only decomposes an already solved problem
  rather than solving it from scratch. SCRL remains effective with the weaker
  generator, improving over GRPO by +2.7 points on average, while DeepSeek-V3.2
  further increases the gain to +3.9 points.

\begin{table*}[t]
  \centering
  \caption{Effect of subproblem generator quality on Qwen3-4B-Base.}
  \label{tab:ablation_generator_main}
  \small
  \setlength{\tabcolsep}{5pt}
  \renewcommand{\arraystretch}{1.10}
  \begin{tabular}{llcccccc}
  \toprule
  \textbf{Method} & \textbf{Subproblem Generator} & \textbf{Olym.B} & \textbf{Minerva} &
  \textbf{MATH} & \textbf{AIME'24} & \textbf{AMC} & \textbf{AVG} \\
  \midrule
  GRPO & \multicolumn{1}{c}{--} & 35.1 & 29.7 & 76.2 & 14.5 & 46.1 & 40.3 \\
  SCRL & Qwen3-4B-Instruct & 36.3 & 34.2 & 80.0 & 16.7 & 47.9 & 43.0 \\
  SCRL & DeepSeek-V3.2 & 39.2 & 33.1 & 79.8 & 16.5 & 52.4 & 44.2 \\
  \bottomrule
  \end{tabular}
  \end{table*}

  Figure~\ref{fig:curriculum_progress_ratio_with_grpo} shows that even with
  DeepSeek-V3.2, the ratio of curriculum instances fully solved at $k_i=4$ remains
  lower than the GRPO solvable ratio (the GRPO bar at $k_i=4$ reports the ratio of
  original problems solved by GRPO under the same half-group counting protocol).
  Nevertheless, SCRL still achieves a substantial performance gain. This indicates
  that SCRL does not require subproblems to be easy or perfectly curated. At the same time, DeepSeek-V3.2
  produces a larger $k_i=4$ ratio than Qwen3-4B-Instruct, suggesting that better
  subproblem quality can further increase SCRL's gains.

\paragraph{\textbf{Subproblem-level normalization enables better credit assignment.}}
  We further ablate on Qwen3-4B-Base how credit is assigned within curriculum rollouts. Table~\ref{tab:ablation_credit_main}
  compares the full method with two alternatives: removing progress-aware
  correction, and Both-GRPO, which keeps mixed training but verifies only
  the final subproblem and applies sample-level GRPO to curriculum rollouts.

  \begin{table*}[t]
  \centering
 \caption{Ablation on credit assignment. Here ``corr'' denotes
    progress-aware correction.}
  \label{tab:ablation_credit_main}
  \small
  \setlength{\tabcolsep}{5pt}
  \renewcommand{\arraystretch}{1.10}
  \begin{tabular}{lcccccc}
  \toprule
  \textbf{Credit Assignment Method} & \textbf{Olym.B} & \textbf{Minerva} & \textbf{MATH} & \textbf{AIME'24} & \textbf{AMC} & \textbf{AVG}
  \\
  \midrule
  Both-GRPO & 38.2 & 33.1 & 79.4 & 14.0 & 54.8 & 43.9 \\
  Subproblem-Level Normalization. w/o corr. & 35.7 & 33.1 & 78.2 & 13.8 & 48.5 & 41.9 \\
  Subproblem-Level Normalization. w/ corr. & 39.2 & 33.1 & 79.8 & 16.5 & 52.4 & 44.2 \\
  \bottomrule
  \end{tabular}
  \end{table*}

Subproblem-level normalization with progress-aware correction performs best.
  Without correction, dense subproblem signals may reward later steps after earlier
  failures, while Both-GRPO uses subproblems only as hints and cannot credit valid
  intermediate progress. Thus, effective curriculum training needs both
  subproblem-specific signals and progress-aware correction.

\FloatBarrier
\section{Conclusion}
\label{sec:conclusion}
We propose SCRL, a subproblem curriculum RL framework for hard LLM reasoning with
  verifiable rewards. SCRL derives verifiable subproblems from reasoning chains and
  uses subproblem-level normalization to convert partial rollout progress into
  token-level learning signals, enabling fine-grained credit assignment without
  external reward models or process annotations. Our theory shows that subproblem
  decomposition can lift hard problems out of gradient dead zones, and experiments
  show consistent gains over strong RLVR and curriculum-learning baselines.
\newpage
\bibliography{main}

\begin{thebibliography}{49}
\providecommand{\natexlab}[1]{#1}
\providecommand{\url}[1]{\texttt{#1}}
\expandafter\ifx\csname urlstyle\endcsname\relax
  \providecommand{\doi}[1]{doi: #1}\else
  \providecommand{\doi}{doi: \begingroup \urlstyle{rm}\Url}\fi

\bibitem[Alam \& Rastogi(2025)Alam and Rastogi]{alam2025limits}
Alam, M.~T. and Rastogi, N.
\newblock Limits of generalization in rlvr: Two case studies in mathematical reasoning.
\newblock \emph{arXiv preprint arXiv:2510.27044}, 2025.

\bibitem[Amani et~al.(2025)Amani, Lotfi, Baldwin, Bengio, Farajtabar, Abbe, and West]{Amani2025RLFR}
Amani, M.~H., Lotfi, A., Baldwin, N.~M., Bengio, S., Farajtabar, M., Abbe, E., and West, R.
\newblock Rl for reasoning by adaptively revealing rationales.
\newblock \emph{ArXiv}, abs/2506.18110, 2025.
\newblock URL \url{https://api.semanticscholar.org/CorpusID:280000657}.

\bibitem[Amari(2016)]{amari2016information}
Amari, S.-i.
\newblock \emph{Information Geometry and Its Applications}, volume 194 of \emph{Applied Mathematical Sciences}.
\newblock Springer Japan, Tokyo, 2016.
\newblock ISBN 978-4-431-55977-1.
\newblock \doi{10.1007/978-4-431-55978-8}.

\bibitem[Bengio et~al.(2009)Bengio, Louradour, Collobert, and Weston]{bengio2009curriculum}
Bengio, Y., Louradour, J., Collobert, R., and Weston, J.
\newblock Curriculum learning.
\newblock In \emph{Proceedings of the 26th annual international conference on machine learning}, pp.\  41--48, 2009.

\bibitem[Chen et~al.(2025)Chen, Peng, Choubey, Huang, Zhang, Bansal, and Wu]{chen2025nudging}
Chen, J. C.-Y., Peng, B.~X., Choubey, P.~K., Huang, K.-H., Zhang, J., Bansal, M., and Wu, C.-S.
\newblock Nudging the boundaries of llm reasoning.
\newblock \emph{arXiv preprint arXiv:2509.25666}, 2025.

\bibitem[Chen et~al.(2026)Chen, Prasad, Khan, Singh, Tian, Stengel-Eskin, and Bansal]{chen2026cog}
Chen, J. C.-Y., Prasad, A., Khan, Z., Singh, J., Tian, R., Stengel-Eskin, E., and Bansal, M.
\newblock Cog-drift: Exploration on adaptively reformulated instances enables learning from hard reasoning problems.
\newblock \emph{arXiv preprint arXiv:2604.04767}, 2026.

\bibitem[Chen(2021)]{chen2021evaluating}
Chen, M.
\newblock Evaluating large language models trained on code.
\newblock \emph{arXiv preprint arXiv:2107.03374}, 2021.

\bibitem[Chu et~al.(2025)Chu, Zhai, Yang, Tong, Xie, Schuurmans, Le, Levine, and Ma]{chu2025sft}
Chu, T., Zhai, Y., Yang, J., Tong, S., Xie, S., Schuurmans, D., Le, Q.~V., Levine, S., and Ma, Y.
\newblock Sft memorizes, rl generalizes: A comparative study of foundation model post-training.
\newblock \emph{arXiv preprint arXiv:2501.17161}, 2025.

\bibitem[Dai et~al.(2026)Dai, Ji, Zhang, Wang, Chu, and Lu]{dai2026harder}
Dai, Y., Ji, Y., Zhang, X., Wang, Y., Chu, X., and Lu, Z.
\newblock Harder is better: Boosting mathematical reasoning via difficulty-aware {GRPO} and multi-aspect question reformulation.
\newblock In \emph{The Fourteenth International Conference on Learning Representations}, 2026.
\newblock URL \url{https://openreview.net/forum?id=nfURupkdRJ}.

\bibitem[Fu et~al.(2025)Fu, Chen, Chai, Wang, Tu, Yin, Lin, Zhang, Zhu, and Zhao]{fu2025srft}
Fu, Y., Chen, T., Chai, J., Wang, X., Tu, S., Yin, G., Lin, W., Zhang, Q., Zhu, Y., and Zhao, D.
\newblock Srft: A single-stage method with supervised and reinforcement fine-tuning for reasoning.
\newblock \emph{arXiv preprint arXiv:2506.19767}, 2025.

\bibitem[Gulcehre et~al.(2023)Gulcehre, Paine, Srinivasan, Konyushkova, Weerts, Sharma, Siddhant, Ahern, Wang, Gu, et~al.]{gulcehre2023reinforced}
Gulcehre, C., Paine, T.~L., Srinivasan, S., Konyushkova, K., Weerts, L., Sharma, A., Siddhant, A., Ahern, A., Wang, M., Gu, C., et~al.
\newblock Reinforced self-training (rest) for language modeling.
\newblock \emph{arXiv preprint arXiv:2308.08998}, 2023.

\bibitem[Guo et~al.(2025)Guo, Yang, Zhang, Song, Zhang, Xu, Zhu, Ma, Wang, Bi, et~al.]{deepseekr1}
Guo, D., Yang, D., Zhang, H., Song, J., Zhang, R., Xu, R., Zhu, Q., Ma, S., Wang, P., Bi, X., et~al.
\newblock Deepseek-r1: Incentivizing reasoning capability in llms via reinforcement learning.
\newblock \emph{arXiv preprint arXiv:2501.12948}, 2025.

\bibitem[Jaech et~al.(2024)Jaech, Kalai, Lerer, Richardson, El-Kishky, Low, Helyar, Madry, Beutel, Carney, et~al.]{jaech2024openai}
Jaech, A., Kalai, A., Lerer, A., Richardson, A., El-Kishky, A., Low, A., Helyar, A., Madry, A., Beutel, A., Carney, A., et~al.
\newblock Openai o1 system card.
\newblock \emph{arXiv preprint arXiv:2412.16720}, 2024.

\bibitem[Li et~al.(2025{\natexlab{a}})Li, Cao, Griggs, Liu, Mo, Tang, Hegde, Hakhamaneshi, Patil, Zaharia, Gonzalez, and Stoica]{Li2025LLMsCE}
Li, D., Cao, S., Griggs, T., Liu, S., Mo, X., Tang, E., Hegde, S., Hakhamaneshi, K., Patil, S.~G., Zaharia, M., Gonzalez, J.~E., and Stoica, I.
\newblock Llms can easily learn to reason from demonstrations structure, not content, is what matters!
\newblock \emph{ArXiv}, abs/2502.07374, 2025{\natexlab{a}}.
\newblock URL \url{https://api.semanticscholar.org/CorpusID:276258697}.

\bibitem[Li et~al.(2026{\natexlab{a}})Li, Lin, Lu, Wen, Yang, Gao, Wu, and Zhang]{li2026questa}
Li, J., Lin, H., Lu, H., Wen, K., Yang, Z., Gao, J., Wu, Y., and Zhang, J.
\newblock Questa: Expanding reasoning capacity in {LLM}s via question augmentation.
\newblock In \emph{The Fourteenth International Conference on Learning Representations}, 2026{\natexlab{a}}.
\newblock URL \url{https://openreview.net/forum?id=3MifB0f7qR}.

\bibitem[Li et~al.(2025{\natexlab{b}})Li, Huang, Wei, Xiong, Wang, and Chu]{Li2025AdaCuRLAC}
Li, R., Huang, H., Wei, F., Xiong, F., Wang, Y., and Chu, X.
\newblock Adacurl: Adaptive curriculum reinforcement learning with invalid sample mitigation and historical revisiting.
\newblock \emph{ArXiv}, abs/2511.09478, 2025{\natexlab{b}}.
\newblock URL \url{https://api.semanticscholar.org/CorpusID:282939669}.

\bibitem[Li et~al.(2026{\natexlab{b}})Li, Chen, Li, Liang, Zhou, Wang, and Zhang]{li2026mathmixup}
Li, X., Chen, J., Li, X., Liang, H., Zhou, X., Wang, T., and Zhang, W.
\newblock Mathmixup: Boosting llm mathematical reasoning with difficulty-controllable data synthesis and curriculum learning.
\newblock \emph{arXiv preprint arXiv:2601.17006}, 2026{\natexlab{b}}.

\bibitem[Liang et~al.(2025)Liang, zhi Li, Gong, Shen, Wu, Guo, and Chen]{Liang2025BeyondPS}
Liang, X., zhi Li, Z., Gong, Y., Shen, Y., Wu, Y., Guo, Z., and Chen, W.
\newblock Beyond pass@1: Self-play with variational problem synthesis sustains rlvr.
\newblock \emph{ArXiv}, abs/2508.14029, 2025.
\newblock URL \url{https://api.semanticscholar.org/CorpusID:280686520}.

\bibitem[Lightman et~al.(2023)Lightman, Kosaraju, Burda, Edwards, Baker, Lee, Leike, Schulman, Sutskever, and Cobbe]{lightman2023let}
Lightman, H., Kosaraju, V., Burda, Y., Edwards, H., Baker, B., Lee, T., Leike, J., Schulman, J., Sutskever, I., and Cobbe, K.
\newblock Let's verify step by step.
\newblock In \emph{The twelfth international conference on learning representations}, 2023.

\bibitem[Lv et~al.(2025)Lv, Zuo, Sun, Liu, Wei, Chen, Zhu, Zhang, Wang, Ding, et~al.]{lv2025towards}
Lv, X., Zuo, Y., Sun, Y., Liu, H., Wei, Y., Chen, Z., Zhu, X., Zhang, K., Wang, B., Ding, N., et~al.
\newblock Towards a unified view of large language model post-training.
\newblock \emph{arXiv preprint arXiv:2509.04419}, 2025.

\bibitem[Ma et~al.(2025)Ma, Liang, Qiang, Tang, Ma, Wong, Niu, Shen, He, Cui, and Zhang]{Ma2025LearningWR}
Ma, L., Liang, H., Qiang, M., Tang, L., Ma, X., Wong, Z.~H., Niu, J., Shen, C., He, R., Cui, B., and Zhang, W.
\newblock Learning what reinforcement learning can't: Interleaved online fine-tuning for hardest questions.
\newblock \emph{ArXiv}, abs/2506.07527, 2025.
\newblock URL \url{https://api.semanticscholar.org/CorpusID:279251208}.

\bibitem[Parashar et~al.(2025)Parashar, Gui, Li, Ling, Vemuri, Olson, Li, Zhang, Caverlee, Kalathil, and Ji]{Parashar2025CurriculumRL}
Parashar, S., Gui, S., Li, X., Ling, H., Vemuri, S., Olson, B., Li, E., Zhang, Y., Caverlee, J., Kalathil, D.~M., and Ji, S.
\newblock Curriculum reinforcement learning from easy to hard tasks improves llm reasoning.
\newblock \emph{ArXiv}, abs/2506.06632, 2025.
\newblock URL \url{https://api.semanticscholar.org/CorpusID:279251658}.

\bibitem[Pikus et~al.(2025)Pikus, Tiwari, and Ye]{Pikus2025HardEA}
Pikus, B., Tiwari, P.~R., and Ye, B.
\newblock Hard examples are all you need: Maximizing grpo post-training under annotation budgets.
\newblock \emph{ArXiv}, abs/2508.14094, 2025.
\newblock URL \url{https://api.semanticscholar.org/CorpusID:280692329}.

\bibitem[Qiyuan et~al.(2026)Qiyuan, Chen, Zhang, and Xu]{qiyuan2026hipo}
Qiyuan, D., Chen, K., Zhang, M., and Xu, Z.
\newblock Hi{PO}: Self-hint policy optimization for {RLVR}.
\newblock In \emph{The Fourteenth International Conference on Learning Representations}, 2026.
\newblock URL \url{https://openreview.net/forum?id=rcb20pHmT1}.

\bibitem[Qu et~al.(2025)Qu, Singh, Lee, Setlur, Salakhutdinov, Finn, and Kumar]{Qu2025RLADTL}
Qu, Y., Singh, A., Lee, Y., Setlur, A.~R., Salakhutdinov, R., Finn, C., and Kumar, A.
\newblock Rlad: Training llms to discover abstractions for solving reasoning problems.
\newblock \emph{ArXiv}, abs/2510.02263, 2025.
\newblock URL \url{https://api.semanticscholar.org/CorpusID:281724383}.

\bibitem[Qu et~al.(2026)Qu, Setlur, Smith, Salakhutdinov, and Kumar]{qu2026pope}
Qu, Y., Setlur, A., Smith, V., Salakhutdinov, R., and Kumar, A.
\newblock Pope: Learning to reason on hard problems via privileged on-policy exploration.
\newblock \emph{arXiv preprint arXiv:2601.18779}, 2026.

\bibitem[Schulman et~al.(2017)Schulman, Wolski, Dhariwal, Radford, and Klimov]{schulman2017proximal}
Schulman, J., Wolski, F., Dhariwal, P., Radford, A., and Klimov, O.
\newblock Proximal policy optimization algorithms.
\newblock \emph{arXiv preprint arXiv:1707.06347}, 2017.

\bibitem[Shao et~al.(2024)Shao, Wang, Zhu, Xu, Song, Bi, Zhang, Zhang, Li, Wu, et~al.]{DeepSeekMath}
Shao, Z., Wang, P., Zhu, Q., Xu, R., Song, J., Bi, X., Zhang, H., Zhang, M., Li, Y., Wu, Y., et~al.
\newblock Deepseekmath: Pushing the limits of mathematical reasoning in open language models.
\newblock \emph{arXiv preprint arXiv:2402.03300}, 2024.

\bibitem[Shenfeld et~al.(2025)Shenfeld, Pari, and Agrawal]{Shenfeld2025RLsRW}
Shenfeld, I., Pari, J., and Agrawal, P.
\newblock Rl's razor: Why online reinforcement learning forgets less.
\newblock \emph{ArXiv}, abs/2509.04259, 2025.
\newblock URL \url{https://api.semanticscholar.org/CorpusID:281103647}.

\bibitem[Sheng et~al.(2025)Sheng, Zhang, Ye, Wu, Zhang, Zhang, Peng, Lin, and Wu]{verl}
Sheng, G., Zhang, C., Ye, Z., Wu, X., Zhang, W., Zhang, R., Peng, Y., Lin, H., and Wu, C.
\newblock Hybridflow: A flexible and efficient rlhf framework.
\newblock In \emph{EuroSys}, pp.\  1279--1297, 2025.
\newblock URL \url{https://doi.org/10.1145/3689031.3696075}.

\bibitem[Shi et~al.(2026)Shi, Chen, Li, Pan, Sun, Xu, Zhou, and Li]{Shi2026R3LRR}
Shi, W., Chen, Y., Li, Z., Pan, X., Sun, Y., Xu, J., Zhou, X., and Li, Y.
\newblock R3l: Reflect-then-retry reinforcement learning with language-guided exploration, pivotal credit, and positive amplification.
\newblock \emph{ArXiv}, abs/2601.03715, 2026.
\newblock URL \url{https://api.semanticscholar.org/CorpusID:284532205}.

\bibitem[Shojaee et~al.(2025)Shojaee, Mirzadeh, Alizadeh, Horton, Bengio, and Farajtabar]{shojaee2025illusion}
Shojaee, P., Mirzadeh, I., Alizadeh, K., Horton, M., Bengio, S., and Farajtabar, M.
\newblock The illusion of thinking: Understanding the strengths and limitations of reasoning models via the lens of problem complexity.
\newblock \emph{arXiv preprint arXiv:2506.06941}, 2025.

\bibitem[Skalse et~al.(2022)Skalse, Howe, Krasheninnikov, and Krueger]{Skalse2022DefiningAC}
Skalse, J. M.~V., Howe, N. H.~R., Krasheninnikov, D., and Krueger, D.
\newblock Defining and characterizing reward hacking.
\newblock \emph{ArXiv}, abs/2209.13085, 2022.
\newblock URL \url{https://api.semanticscholar.org/CorpusID:252545256}.

\bibitem[Trinh et~al.(2024)Trinh, Wu, Le, He, and Luong]{trinh2024solving}
Trinh, T.~H., Wu, Y., Le, Q.~V., He, H., and Luong, T.
\newblock Solving olympiad geometry without human demonstrations.
\newblock \emph{Nature}, 625\penalty0 (7995):\penalty0 476--482, 2024.

\bibitem[Uesato et~al.(2022)Uesato, Kushman, Kumar, Song, Siegel, Wang, Creswell, Irving, and Higgins]{uesato2022solving}
Uesato, J., Kushman, N., Kumar, R., Song, F., Siegel, N., Wang, L., Creswell, A., Irving, G., and Higgins, I.
\newblock Solving math word problems with process-and outcome-based feedback.
\newblock \emph{arXiv preprint arXiv:2211.14275}, 2022.

\bibitem[Wang et~al.(2025)Wang, Yu, Gao, Zheng, Liu, Lu, Dang, Chen, Yang, Zhang, et~al.]{wang2025beyond}
Wang, S., Yu, L., Gao, C., Zheng, C., Liu, S., Lu, R., Dang, K., Chen, X., Yang, J., Zhang, Z., et~al.
\newblock Beyond the 80/20 rule: High-entropy minority tokens drive effective reinforcement learning for llm reasoning.
\newblock \emph{arXiv preprint arXiv:2506.01939}, 2025.

\bibitem[Wen et~al.(2025)Wen, Liu, Zheng, Xu, Ye, Wu, Liang, Wang, Li, Miao, Bian, and Yang]{Wen2025ReinforcementLW}
Wen, X., Liu, Z., Zheng, S., Xu, Z., Ye, S., Wu, Z., Liang, X., Wang, Y., Li, J., Miao, Z., Bian, J., and Yang, M.
\newblock Reinforcement learning with verifiable rewards implicitly incentivizes correct reasoning in base llms.
\newblock \emph{ArXiv}, abs/2506.14245, 2025.
\newblock URL \url{https://api.semanticscholar.org/CorpusID:279410727}.

\bibitem[Wu et~al.(2025{\natexlab{a}})Wu, Liao, Feng, Zhang, Wen, Shao, Xu, and Tao]{wu2025thought}
Wu, J., Liao, C., Feng, M., Zhang, S., Wen, Z., Shao, P., Xu, H., and Tao, J.
\newblock Thought-augmented policy optimization: Bridging external guidance and internal capabilities.
\newblock \emph{arXiv preprint arXiv:2505.15692}, 1\penalty0 (8):\penalty0 10, 2025{\natexlab{a}}.

\bibitem[Wu et~al.(2025{\natexlab{b}})Wu, Qian, Liu, Wang, Huang, Liang, Miao, Dou, Lv, Wang, Xu, Chen, Li, Zheng, and Huang]{Wu2025ProgressiveMC}
Wu, M., Qian, Q., Liu, W., Wang, X., Huang, Z., Liang, D., Miao, L., Dou, S., Lv, C., Wang, Z., Xu, Z., Chen, L., Li, T., Zheng, X., and Huang, X.
\newblock Progressive mastery: Customized curriculum learning with guided prompting for mathematical reasoning.
\newblock \emph{ArXiv}, abs/2506.04065, 2025{\natexlab{b}}.
\newblock URL \url{https://api.semanticscholar.org/CorpusID:279154725}.

\bibitem[Yan et~al.(2025)Yan, Li, Hu, Wang, Cui, Qu, Cheng, and Zhang]{Luffy}
Yan, J., Li, Y., Hu, Z., Wang, Z., Cui, G., Qu, X., Cheng, Y., and Zhang, Y.
\newblock Learning to reason under off-policy guidance.
\newblock In \emph{The Thirty-ninth Annual Conference on Neural Information Processing Systems}, 2025.
\newblock URL \url{https://openreview.net/forum?id=vO8LLoNWWk}.

\bibitem[Yang et~al.(2024)Yang, Zhang, Hui, Gao, Yu, Li, Liu, Tu, Zhou, Lin, et~al.]{yang2024qwen2}
Yang, A., Zhang, B., Hui, B., Gao, B., Yu, B., Li, C., Liu, D., Tu, J., Zhou, J., Lin, J., et~al.
\newblock Qwen2. 5-math technical report: Toward mathematical expert model via self-improvement.
\newblock \emph{arXiv preprint arXiv:2409.12122}, 2024.

\bibitem[Yang et~al.(2025)Yang, Wu, Liu, Zhang, Li, Liang, Wang, Nie, Xu, Shi, Huang, and Zhang]{Yang2025FromIT}
Yang, C., Wu, J., Liu, Y., Zhang, S., Li, Y., Liang, Q., Wang, H., Nie, S., Xu, J., Shi, R., Huang, Y., and Zhang, G.
\newblock From imitation to discrimination: Toward a generalized curriculum advantage mechanism enhancing cross-domain reasoning tasks.
\newblock In \emph{AAAI Conference on Artificial Intelligence}, 2025.
\newblock URL \url{https://api.semanticscholar.org/CorpusID:283458223}.

\bibitem[Yang et~al.(2026)Yang, Bai, Wu, Yang, Setlur, and Kumar]{yang2026int}
Yang, M.~Y., Bai, H., Wu, I., Yang, G., Setlur, A., and Kumar, A.
\newblock Int: Self-proposed interventions enable credit assignment in llm reasoning.
\newblock \emph{arXiv preprint arXiv:2601.14209}, 2026.

\bibitem[Yu et~al.(2025)Yu, Zhang, Zhu, Yuan, Zuo, Yue, Fan, Liu, Liu, Liu, Lin, Lin, Ma, Sheng, Tong, Zhang, Zhang, Zhang, Zhu, Zhu, Chen, Chen, Wang, Yu, Dai, Song, Wei, Zhou, Liu, Ma, Zhang, Yan, Qiao, Wu, and Wang]{Yu2025DAPOAO}
Yu, Q., Zhang, Z., Zhu, R., Yuan, Y., Zuo, X., Yue, Y., Fan, T., Liu, G., Liu, L., Liu, X., Lin, H., Lin, Z., Ma, B., Sheng, G., Tong, Y., Zhang, C., Zhang, M., Zhang, W., Zhu, H., Zhu, J., Chen, J., Chen, J., Wang, C., Yu, H., Dai, W., Song, Y., Wei, X., Zhou, H., Liu, J., Ma, W., Zhang, Y.-Q., Yan, L., Qiao, M., Wu, Y.-X., and Wang, M.
\newblock Dapo: An open-source llm reinforcement learning system at scale.
\newblock \emph{ArXiv}, abs/2503.14476, 2025.
\newblock URL \url{https://api.semanticscholar.org/CorpusID:277104124}.

\bibitem[Yue et~al.(2025)Yue, Chen, Lu, Zhao, Wang, Song, and Huang]{yue2025does}
Yue, Y., Chen, Z., Lu, R., Zhao, A., Wang, Z., Song, S., and Huang, G.
\newblock Does reinforcement learning really incentivize reasoning capacity in llms beyond the base model?
\newblock \emph{arXiv preprint arXiv:2504.13837}, 2025.

\bibitem[Zelikman et~al.(2022)Zelikman, Wu, and Goodman]{zelikman2022star}
Zelikman, E., Wu, Y., and Goodman, N.~D.
\newblock Star: Self-taught reasoner.
\newblock \emph{arXiv preprint arXiv:2203.14465}, 2022.

\bibitem[Zhang et~al.(2024)Zhang, Zhoubian, Hu, Yue, Dong, and Tang]{zhang2024rest}
Zhang, D., Zhoubian, S., Hu, Z., Yue, Y., Dong, Y., and Tang, J.
\newblock Rest-mcts*: Llm self-training via process reward guided tree search.
\newblock \emph{Advances in Neural Information Processing Systems}, 37:\penalty0 64735--64772, 2024.

\bibitem[Zhang et~al.(2025{\natexlab{a}})Zhang, Lv, Li, Wang, Wang, Hu, and Yan]{zhang2025stephint}
Zhang, K., Lv, A., Li, J., Wang, Y., Wang, F., Hu, H., and Yan, R.
\newblock Stephint: Multi-level stepwise hints enhance reinforcement learning to reason.
\newblock \emph{arXiv preprint arXiv:2507.02841}, 2025{\natexlab{a}}.

\bibitem[Zhang et~al.(2025{\natexlab{b}})Zhang, Wu, Zhu, Tan, Yu, He, and Jia]{zhang2025scaf}
Zhang, X., Wu, S., Zhu, Y., Tan, H., Yu, S., He, Z., and Jia, J.
\newblock Scaf-grpo: Scaffolded group relative policy optimization for enhancing llm reasoning.
\newblock \emph{arXiv preprint arXiv:2510.19807}, 2025{\natexlab{b}}.

\end{thebibliography}
\bibliographystyle{icml2025}

\appendix
\newpage
\section{More Ablation Study}
\label{sec:ablation}

\paragraph{Number of subproblems.}
  We first ablate the number of subproblems $K$.
  Our main setting uses $K=4$, with the final subproblem fixed as the original
  problem.
  For $K=3$ and $K=2$, we take the last three or last two subproblems from the
  $K=4$ sequence, so the shorter curricula preserve the nested structure and still
  end with the original problem.

  \begin{table*}[h]
  \centering
  \caption{Ablation on the number of subproblems $K$.}
  \label{tab:ablation_scaffold_length}
  \resizebox{\textwidth}{!}{
  \begin{tabular}{lcccccc}
  \toprule
  Methods & AIME24 & AMC & Minerva & OlympiadBench & MATH & Avg. \\
  \midrule
  GRPO & 14.5 & 46.1 & 29.7 & 35.1 & 76.2 & 40.3 \\
  SCRL ($K=2$) & 14.0 & 49.7 & 30.1 & 36.1 & 79.2 & 41.8 \\
  SCRL ($K=3$) & 17.0 & 51.8 & 33.8 & 35.1 & 76.2 & 42.8 \\
  SCRL ($K=4$) & 16.5 & 52.4 & 33.1 & 39.2 & 79.8 & 44.2 \\
  \bottomrule
  \end{tabular}
  }
  \end{table*}

\begin{figure}[h] 
    \centering
    \includegraphics[width=0.8\linewidth]{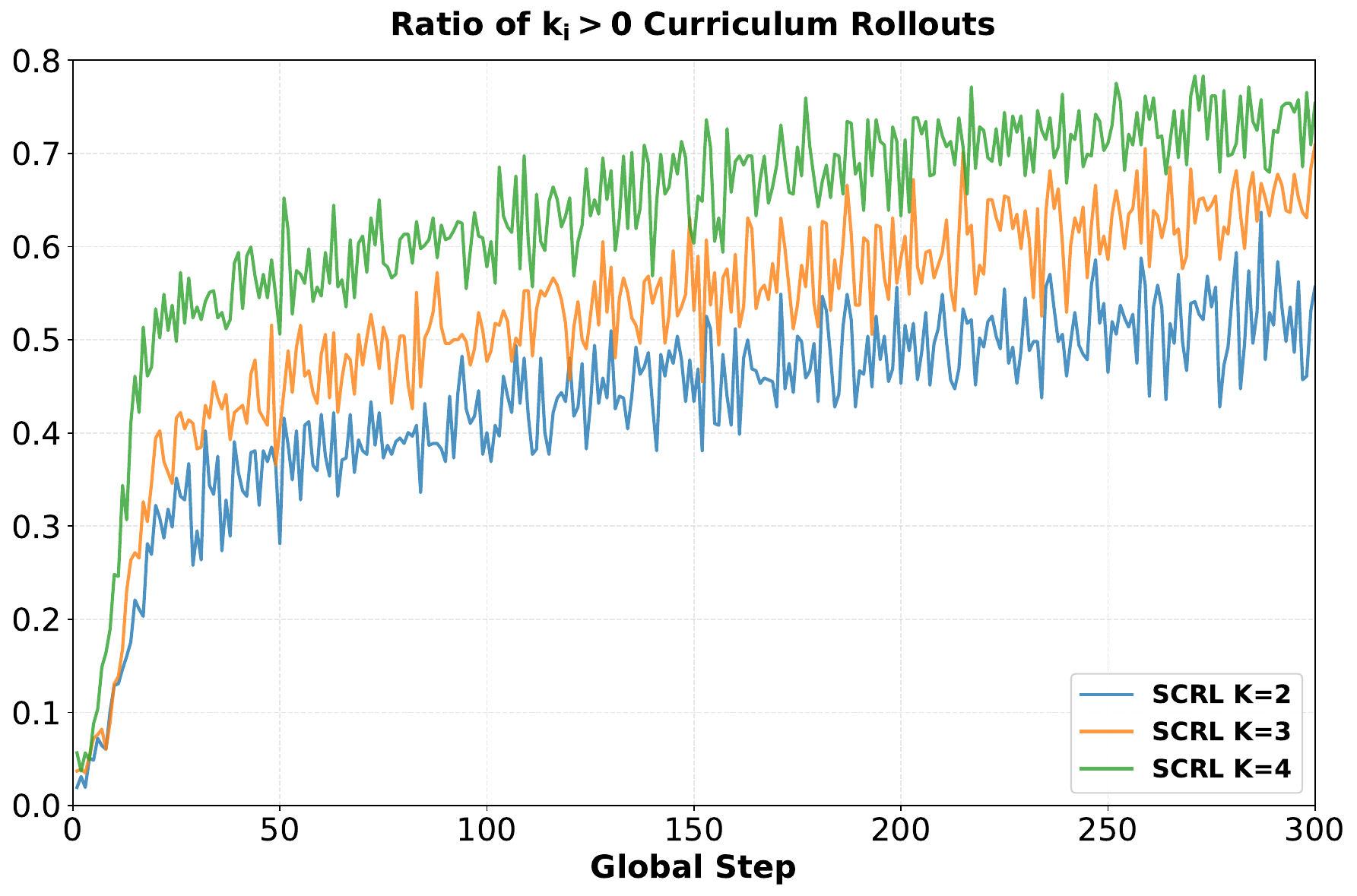}
    \caption{Number of $k_i>0$ curriculum rollouts during training.}
    \label{fig:ablation_zero_progress}
\end{figure}

  Table~\ref{tab:ablation_scaffold_length} shows that increasing $K$ improves the
  average performance, with $K=4$ performing best.
  This supports the role of subproblem curricula: more subproblems expose a finer
  progression toward the original problem, creating more verifiable intermediate
  signals within each rollout. Figure~\ref{fig:ablation_zero_progress} explains why longer subproblem
  curricula help.
  As $K$ increases, fewer rollouts make zero progress, meaning more rollouts solve
  at least one subproblem and receive a non-empty learning signal.
  Thus, increasing $K$ improves both the granularity of credit assignment and the
  availability of reward signal on hard problems.

  However, longer curricula also increase rollout complexity.
  The model must answer more subproblems, and progress-aware correction requires
  earlier subproblems to be solved before later ones can receive credit.
  If an intermediate subproblem is ambiguous or poorly constructed, it can block
  credit for later progress.
  We therefore use $K=4$ as a practical trade-off between denser supervision and
  curriculum complexity.

 \paragraph{Training data construction.}
  Each SCRL curriculum rollout contains $K=4$ subproblems, so we ask whether the
  gain comes from the training algorithm or simply from exposing the model to more
  questions. We compare the default setting, \texttt{hard\_1024} with SCRL, against
  two data-scaling controls. \texttt{hard\_4096} expands the original-problem set by
  adding $3\times1024$ hard problems from the InT dataset~\citep{yang2026int}.
  \texttt{subproblem\_4096} instead splits each four-subproblem curriculum instance
  into four standalone questions; since the last subproblem is the original
  problem, it also contains \texttt{hard\_1024}.

  \begin{table*}[t]
  \centering
  \caption{Ablation on training data construction. Avg. Resp. Len. is measured
  during mixed training and averages over both original-problem rollouts and
  curriculum rollouts.}
  \label{tab:ablation_data}
  \resizebox{\textwidth}{!}{
  \begin{tabular}{llccccccc}
  \toprule
  Dataset & Training Objective & AIME24 & AMC & Minerva & OlympiadBench & MATH & Avg. & Avg. Resp. Len. \\
  \midrule
  \texttt{hard\_1024} + subproblem bank & SCRL & 16.5 & 52.4 & 33.1 & 39.2 & 79.8 & 44.2 & 1752 \\
  \texttt{hard\_1024} & GRPO & 14.5 & 46.1 & 29.7 & 35.1 & 76.2 & 40.3 & 1161 \\
  \texttt{hard\_4096} & GRPO & 14.2 & 47.6 & 31.3 & 35.5 & 77.6 & 41.2 & 1256 \\
  \texttt{subproblem\_4096} & GRPO & 17.1 & 44.9 & 31.6 & 37.0 & 77.8 & 41.7 & 905 \\
  \bottomrule
  \end{tabular}
  }
  \end{table*}

Table~\ref{tab:ablation_data} shows that SCRL outperforms both data-scaling
  controls. This suggests that the gain does not mainly come from seeing more
  questions, but from using subproblems as a curriculum inside each rollout.
  Subproblems serve as intermediate anchors: after solving an earlier subproblem,
  the model can build on that result when solving later ones, while
  subproblem-level normalization assigns credit to the corresponding answer spans.
  The average response length is computed over the first 20 training epochs.
  Although each SCRL curriculum rollout contains four subproblems, its average
  response length is only about $1.5\times$ that of GRPO on \texttt{hard\_1024}.
  This indicates that SCRL does not spend response length proportional to the
  number of subproblems, but instead uses the subproblem curriculum structure to
  support more efficient exploration.

The two controls further clarify this point. \texttt{hard\_4096} improves over
  \texttt{hard\_1024}, but the gain is limited, showing that simply adding more
  hard problems is less effective under our training budget.
  \texttt{subproblem\_4096} brings a slightly larger gain, but still falls behind
  SCRL, suggesting that training on isolated easier subproblems does not by itself
  teach the model to solve harder target problems. In contrast, SCRL keeps the
  model exploring near its current capability boundary by preserving the
  dependency among subproblems, making the curriculum more useful than either data
  scaling strategy.

\section{Proofs for Section~\ref{sec:theory}}
\label{app:proofs}

\subsection{Proof of Theorem~\ref{thm:dead_zone} (\texorpdfstring{Bound on $\lambda_{\min}(\bm{F}_x(\theta))$}{Bound on Lambda\_min})}

We bound $\lambda_{\min}(\bm{F}_x(\theta))$ when $p(x;\theta) < \delta$.

Let $E_0$ denote the event that all $G$ rollouts share the same reward
(either all fail or all succeed).
On $E_0$, the group is degenerate ($\hat{\sigma}_{r_x} = 0$),
so by the GRPO convention $\hat{A}_i = 0$ for all $i$,
and every term in~\eqref{eq:egim} is zero.
By the law of total expectation:
\begin{equation}
    \bm{F}_x(\theta)
    \;=\; \Pr[E_0^c]\cdot
    \mathbb{E}\!\left[
    \frac{1}{G}\sum_i\hat{A}_i^2 s_i s_i^\top \,\Big|\, E_0^c
    \right],
    \label{eq:app_total_exp}
\end{equation}
where $s_i = \nabla_\theta\log\pi_\theta(o_i \mid x)$.
For any unit vector $v$, using $\hat{A}_i^2 \leq C_{\hat{A}}^2$ a.s.\ and
$(v^\top s_i)^2 \leq B_s^2$ a.s.\ (by regularity):
\begin{equation}
    v^\top\bm{F}_x(\theta)\,v
    \;\leq\; \Pr[E_0^c]\cdot C_{\hat{A}}^2 \cdot B_s^2.
    \label{eq:app_quadratic}
\end{equation}
Since $\Pr[E_0^c] = 1 - p^G - (1-p)^G \leq 1-(1-p)^G \leq 1-(1-\delta)^G \leq G\delta$
(using $p < \delta$ and Bernoulli's inequality):
\begin{equation}
    \lambda_{\min}\!\left(\bm{F}_x(\theta)\right)
    \;\leq\;
    \sup_{\|v\|=1} v^\top\bm{F}_x v
    \;\leq\; G\delta\cdot C_{\hat{A}}^2\cdot B_s^2
    \;=\; O(\delta).
    \label{eq:app_dead_final}
\end{equation}
This establishes Theorem~\ref{thm:dead_zone}.\qed

  \subsection{Proof of Theorem~\ref{thm:recovery}}

  \paragraph{First claim.}
  We bound the column-$1$ contribution to $\bm{F}_{\mathcal{T}(x)}(\theta)$.
  Let $E_1^{\neq}$ denote the event that column~1 is non-degenerate.

  \textit{Lower bound on $\Pr[E_1^{\neq}]$.}
  Since $p_1 \in [p^\star, 1-p^\star]$ by assumption, and $g(p) = 1-p^G-(1-p)^G$
  is symmetric around $1/2$ and non-decreasing on $[0, 1/2]$, its minimum on
  $[p^\star, 1-p^\star]$ is attained at the endpoints:
  \begin{equation}
      \Pr[E_1^{\neq}]
      \;=\; g(p_1)
      \;\geq\; g(p^\star)
      \;=\; 1 - (p^\star)^G - (1-p^\star)^G
      \;=:\; q_{\min} \;>\; 0.
      \label{eq:app_q_bound}
  \end{equation}

  \textit{Lower bound on the conditional EGIM.}
  We compute $v^\top\bm{F}^{(1)}_{\mathcal{T}(x)}v$ by conditioning on the
  reward vector $(R_1^{(1)},\ldots,R_G^{(1)})$.
  Given the rewards, $\hat{A}_i^{(1)}$ is fully determined, and since rollouts
  are i.i.d., $o_i^{(1)}$ is conditionally independent of $(R_j^{(1)})_{j\neq i}$
  given $R_i^{(1)}$, so
  $\mathbb{E}[(v^\top s_i^{(1)})^2\mid (R_1^{(1)},\ldots,R_G^{(1)})]
  = \mathbb{E}[(v^\top s_i^{(1)})^2\mid R_i^{(1)}]$.
  By the tower property:
  \begin{align}
      v^\top\bm{F}^{(1)}_{\mathcal{T}(x)}v
      &\;=\; \frac{1}{G}\sum_{i=1}^G
      \mathbb{E}\!\left[
          \hat{A}_i^{(1)2}\,
          \mathbb{E}\!\left[(v^\top s_i^{(1)})^2 \mid R_i^{(1)}\right]
      \right] \notag \\
      &\;\geq\; \sigma_{\min}^2 \cdot
      \frac{1}{G}\sum_{i=1}^G \mathbb{E}\!\left[\hat{A}_i^{(1)2}\right],
      \label{eq:app_cond_egim}
  \end{align}
  where the inequality uses the conditional identifiability assumption
($\mathbb{E}[(v^\top s)^2 \mid r = r_i^{(1)}] \geq \sigma_{\min}^2$
for all $r_i^{(1)} \in \{0,1\}$) and $\hat{A}_i^{(1)2} \geq 0$.
By definition of $\hat{\sigma}_{r}$, the sample average of squared
advantages satisfies
\begin{equation}
    \frac{1}{G}\sum_{i=1}^{G}\hat{A}_i^{(1)2} \;=\; 1
    \quad\text{on } E_1^{\neq}
    \quad\text{and}\quad
    = 0 \text{ otherwise,}
    \label{eq:adv_norm}
\end{equation}
so $\frac{1}{G}\sum_i\mathbb{E}[\hat{A}_i^{(1)2}] = \Pr[E_1^{\neq}] \geq q_{\min}$.
Hence:

  \textit{Conclusion:}
  Substituting into~\eqref{eq:app_cond_egim}:
  \begin{equation}
      v^\top\bm{F}^{(1)}_{\mathcal{T}(x)}\,v
      \;\geq\; \sigma_{\min}^2 \cdot \Pr[E_1^{\neq}]
      \;\geq\; q_{\min}\cdot\sigma_{\min}^2.
      \label{eq:app_psd_lb}
  \end{equation}
  Since all terms in~\eqref{eq:lifted_egim} are PSD,
  $\bm{F}_{\mathcal{T}(x)} \succeq \tfrac{1}{K}\bm{F}^{(1)}_{\mathcal{T}(x)}$, hence:
  \begin{equation}
      \lambda_{\min}\!\left(\bm{F}_{\mathcal{T}(x)}(\theta)\right)
      \;\geq\; \tfrac{1}{K}\cdot q_{\min}\cdot\sigma_{\min}^2
      \;=: \tfrac{1}{K}\cdot c(p^\star, G, \sigma_{\min})
      \;>\; 0.
      \label{eq:app_recovery}
  \end{equation}

  \paragraph{Second claim.}
  Combining~\eqref{eq:app_dead_final} and~\eqref{eq:app_recovery}:
  \begin{equation}
      \frac{\lambda_{\min}(\bm{F}_{\mathcal{T}(x)}(\theta))}
           {\lambda_{\min}(\bm{F}_x(\theta))}
      \;\geq\;
      \frac{\tfrac{1}{K}\cdot c(p^\star, G, \sigma_{\min})}
           {G\delta\cdot C_{\hat{A}}^2\cdot B_s^2}
      \;=\; \Omega\!\left(\frac{1}{\delta}\right),
      \label{eq:app_amplification}
  \end{equation}
  since the numerator is independent of $\delta$.\qed

\subsection{Bound on $C_{\hat{A}}$}
\label{app:adv_bound}

We show that for binary rewards $r_i \in \{0,1\}$, the group-normalized GRPO
  advantage satisfies $C_{\hat{A}} := \sup_i |\hat{A}_i| \leq \sqrt{G-1}$.
For a non-degenerate group with $k$ successes ($1 \leq k \leq G-1$),
$\bar{r} = k/G$ and $\hat{\sigma}_r = \sqrt{(k/G)(1-k/G)}$.
The advantage of a success rollout is:
\begin{equation}
    |\hat{A}_{\text{success}}|
    \;=\; \frac{1-k/G}{\sqrt{(k/G)(1-k/G)}}
    \;=\; \sqrt{\frac{1-k/G}{k/G}}
    \;=\; \sqrt{\frac{G-k}{k}},
    \label{eq:app_adv_success}
\end{equation}
which is maximized at $k=1$, giving $|\hat{A}| = \sqrt{G-1}$.
By symmetry, the advantage of a failure rollout is $\sqrt{k/(G-k)}$,
also maximized at $k=G-1$, giving $\sqrt{G-1}$.
Hence $C_{\hat{A}} = \sqrt{G-1} < \sqrt{G}$.\qed

\section{SCRL Training Algorithm}
\label{app:scrl_algorithm}

\begin{algorithm}[H]
\caption{SCRL Training}
\label{alg:scrl_training}
\begin{algorithmic}[1]
\REQUIRE Problem set $\mathcal{D}$, policy $\pi_\theta$, subproblem bank
$\{\mathbf{s}(x)\}_{x \in \mathcal{D}}$, subproblem count $K$, group size $G$
\FOR{each training step}
    \STATE Sample $x \sim \mathcal{D}$ and form the curriculum prompt $t_K(x)$
    \STATE Sample $G/2$ curriculum rollouts
           $\{o_i\}_{i=1}^{G/2}$ from $\pi_\theta(\cdot \mid t_K(x))$
    \FOR{each curriculum rollout $o_i$, $i=1,\ldots,G/2$}
        \STATE Extract $K$ answer spans using the required response format
        \STATE Verify each subproblem answer to obtain raw reward vector
               $\mathbf{r}_i$
        \STATE Compute curriculum progress $k_i$ using Eq.~\eqref{eq:progress}
        \STATE Apply progress-aware correction to obtain
               $\tilde{\mathbf{r}}_i$ using Eq.~\eqref{eq:pca}
        \STATE Set final subproblem rewards
               $R_i^{(j)} := \tilde{r}_i^{(j)}$ for $j=1,\ldots,K$
    \ENDFOR
    \FOR{each subproblem position $j = 1,\ldots,K$}
        \STATE Form $\mathbf{R}^{(j)}=(R_1^{(j)},\ldots,R_{G/2}^{(j)})$
        \STATE Compute subproblem-level advantages $A_i^{(j)}$ for
               $i=1,\ldots,G/2$ using Eq.~\eqref{eq:col_grpo}
    \ENDFOR
    \STATE Assign token-level advantages $A_{i,t}=A_i^{(\mathrm{sub}_i(t))}$
           to curriculum rollout tokens according to their tagged answer spans
    \STATE Sample $G/2$ original-problem rollouts
           $\{o_i\}_{i=G/2+1}^{G}$ from $\pi_\theta(\cdot \mid x)$
    \STATE Verify final answers and compute rollout-level advantages $A_i$
           for $i=G/2+1,\ldots,G$ using standard GRPO
    \STATE Update $\pi_\theta$ using the SCRL objective in
           Eq.~\eqref{eq:loss_scrl_full}
\ENDFOR
\end{algorithmic}
\end{algorithm}

  \section{OOD Task Performance}
  \label{app:ood_performance}

  \begin{table}[h]
  \centering
  \caption{Out-of-distribution evaluation on Qwen3-14B-Base.}
  \label{tab:ood_14b}
  \begin{tabular}{lcccc}
  \toprule
  Method & GPQA & HumanEval & LiveCodeBench v6 & Avg. Score \\
  \midrule
  Base Model & 38.89 & 82.93 & 19.80 & 47.20 \\
  GRPO & 36.86 & 84.15 & 24.10 & 48.37 \\
  \textbf{SCRL (Ours)} & \textbf{41.41} & \textbf{89.02} & \textbf{24.57} & \textbf{51.67} \\
  \bottomrule
  \end{tabular}
  \end{table}

  \paragraph{\textbf{SCRL generalizes to out-of-distribution tasks.}}
  To examine whether the gains from SCRL transfer beyond the mathematical
  benchmarks used for training, we evaluate the Qwen3-14B-Base model on three
  out-of-distribution benchmarks: GPQA, HumanEval, and LiveCodeBench v6. These
  benchmarks cover different reasoning domains, including scientific question
  answering and code generation, and are not used for constructing the subproblem
  curriculum.

  As shown in Table~\ref{tab:ood_14b}, SCRL achieves the best average OOD score,
  reaching 51.67 compared with 47.20 for the base model and 48.37 for GRPO. SCRL
  also improves consistently across all three OOD benchmarks, with gains on GPQA
  (41.41 vs. 38.89 for the base model and 36.86 for GRPO), HumanEval (89.02 vs.
  82.93 and 84.15), and LiveCodeBench v6 (24.57 vs. 19.80 and 24.10). These results
  suggest that SCRL does not merely overfit to the generated curriculum prompts or
  the training benchmark distribution. Instead, the subproblem curriculum appears
  to improve transferable reasoning behavior, including domains where solutions
  require multi-step reasoning or program synthesis rather than the exact
  mathematical format used during training.

\section{Detailed Experimental Results}
\label{app:full_results}

Here we provide the complete Pass@$k$ performance ($k \in \{1, 2, 4, 8, 16, 32, 64\}$) for Qwen3-4B-Base.
\begin{table}[H]
    \centering
    \caption{Full results on \textbf{Qwen3-4B-Base}.}
    \label{tab:full_olympiad}
    \resizebox{1.0\textwidth}{!}{
    \begin{tabular}{l|l|ccccccc}
        \toprule
        \textbf{Base Model} & \textbf{Method} & \textbf{P@1} & \textbf{P@2} & \textbf{P@4} & \textbf{P@8} & \textbf{P@16} & \textbf{P@32} & \textbf{P@64} \\
        \midrule
        \multirow{7}{*}{OlympiadBench} & Initial Model & 29.0 & 37.2 & 43.8 & 49.5 & 54.5 & 58.6 & 62.3 \\
         & SFT & 31.1 & 36.4 & 43.4 & 49.7 & 55.0 & \underline{59.5} & \textbf{63.4} \\
        & GRPO & 35.1 & 40.0 & 44.0 & 47.4 & 50.5 & 53.5 & 56.4 \\
        & DAPO & 34.9 & \underline{41.3} & \underline{46.9} & \underline{51.7} & \underline{55.8} & 59.1 & 61.6 \\
        & QuestA & 35.0 & 40.0 & 44.4 & 48.1 & 51.3 & 54.1 & 56.6 \\
        & NuRL & \underline{35.6} & 40.9 & 45.4 & 49.3 & 52.6 & 55.5 & 58.2 \\
        & \textbf{SCRL (Ours)} & \textbf{39.2} & \textbf{45.2} & \textbf{50.4} & \textbf{54.7} & \textbf{58.1} & \textbf{60.9} & \underline{63.0} \\
        \midrule
        \multirow{7}{*}{Minerva} & Initial Model & 15.6 & 22.4 & 29.6 & 36.8 & 43.5 & 48.9 & \underline{52.9} \\
             & SFT & 25.4 & 27.6 & 30.8 & 36.7 & 41.4 & 45.5 & 49.6\\
        & GRPO & 29.7 & 33.9 & 37.5 & 40.6 & 43.6 & 46.3 & 48.2 \\
        & DAPO & 30.1 & \underline{35.7} & \underline{40.1} & \underline{43.7} & \underline{47.1} & \underline{50.1} & 52.6 \\
        & QuestA & \underline{32.0} & 35.6 & 38.8 & 41.6 & 43.9 & 46.0 & 48.2 \\
        & NuRL & 30.3 & 34.7 & 38.4 & 41.8 & 45.1 & 48.2 & 51.1 \\
        & \textbf{SCRL (Ours)} & \textbf{33.1} & \textbf{37.5} & \textbf{41.3} & \textbf{44.8} & \textbf{48.2} & \textbf{51.6} & \textbf{54.8} \\
        \midrule
        \multirow{7}{*}{MATH-500} & Initial Model & 64.4 & 76.7 & 83.6 & 87.7 & 90.6 & 92.8 & 94.2 \\
             & SFT & 71.0 & 73.3 & 83.2 & \underline{89.0} & \textbf{92.5} & \textbf{94.8} & \textbf{96.4} \\
        & GRPO & 76.2 & 80.5 & 83.9 & 86.5 & 88.6 & 90.3 & 91.6 \\
        & DAPO & 76.0 & \underline{82.1} & \underline{86.3} & 89.4 & \underline{91.6} & 93.1 & 93.8 \\
        & QuestA & 75.9 & 80.3 & 83.8 & 86.4 & 88.3 & 89.7 & 90.6 \\
        & NuRL & \underline{77.6} & \underline{82.1} & 85.2 & 87.4 & 89.2 & 90.7 & 91.8 \\
        & \textbf{SCRL (Ours)} & \textbf{79.8} & \textbf{84.6} & \textbf{87.8} & \textbf{90.0} & \underline{91.6} & \underline{93.2} & \underline{94.8} \\
        \midrule        
        \multirow{7}{*}{AIME2024} & Initial Model & 10.9 & 16.4 & 22.0 & 27.3 & 32.4 & 37.6 & 43.3 \\
             & SFT & 10.6 & 15.7 & 21.9 & \underline{28.9} & \underline{36.1} & \underline{43.3} & \underline{50.0} \\
        & GRPO & \underline{14.5} & \underline{19.1} & \underline{23.4} & 27.4 & 31.3 & 34.9 & 40.0 \\
        & DAPO & 11.8 & 16.5 & 21.6 & 27.7 & 35.7 & \textbf{45.7} & \textbf{56.7} \\
        & QuestA & 14.1 & 17.6 & 20.8 & 23.6 & 27.3 & 32.9 & 40.0 \\
        & NuRL & 13.2 & 18.0 & 22.8 & 27.9 & 33.9 & 40.3 & 46.7 \\
        & \textbf{SCRL (Ours)} & \textbf{16.5} & \textbf{22.1} & \textbf{27.4} & \textbf{33.0} & \textbf{38.2} & 42.4 & 46.7 \\
        \midrule
        \multirow{7}{*}{AIME2025} & Initial Model & 6.4 & 10.8 & 16.4 & 22.1 & 27.9 & 35.4 & \underline{46.7} \\
             & SFT & 10.0 & 14.6 & \underline{22.0} & \textbf{30.1} & \textbf{38.4} & \textbf{47.3} & \textbf{56.7} \\
        & GRPO & 7.2 & 12.0 & 18.0 & 24.5 & 31.1 & 37.7 & 43.3 \\
        & DAPO & 9.3 & 13.9 & 18.5 & 23.3 & 29.3 & 36.4 & 43.3 \\
        & QuestA & \underline{11.7} & \underline{16.0} & 20.2 & 24.5 & 30.3 & 37.3 & 43.3 \\
        & NuRL & 8.4 & 12.9 & 17.7 & 22.5 & 27.2 & 32.5 & 40.0 \\
        & \textbf{SCRL (Ours)} & \textbf{15.3} & \textbf{19.3} & \textbf{23.0} & \underline{26.9} & \underline{32.1} & \underline{38.8} & \underline{46.7} \\
        \midrule
        \multirow{7}{*}{AMC} & Initial Model & 38.1 & 50.3 & 60.5 & 69.0 & 76.1 & 82.5 & \underline{89.2} \\
             & SFT & 41.0 & 47.7 & 59.2 & \underline{69.4} & \underline{78.0} & \textbf{85.2} & \textbf{90.4} \\
        & GRPO & 46.1 & 54.0 & 61.6 & 68.6 & 74.8 & 79.7 & 84.3 \\
        & DAPO & \underline{46.8} & \underline{56.2} & \underline{64.5} & 71.3 & 76.7 & 81.9 & 86.7 \\
        & QuestA & \underline{46.8} & 53.6 & 60.0 & 66.1 & 71.9 & 77.7 & 84.3 \\
        & NuRL & 45.4 & 53.3 & 61.0 & 68.5 & 75.2 & 81.0 & 85.5 \\
        & \textbf{SCRL (Ours)} & \textbf{52.4} & \textbf{61.0} & \textbf{68.6} & \textbf{74.8} & \textbf{80.4} & \underline{85.0} & 88.0 \\
        \midrule
        \multirow{7}{*}{IMO-Bench} & Initial Model & 6.1 & 9.7 & 14.0 & 18.5 & 22.9 & 27.3 & \underline{31.6} \\
             & SFT & 7.1 & 11.3 & \underline{16.4} & \textbf{21.9} & \textbf{27.4} & \textbf{32.8} & \textbf{39.0} \\
        & GRPO & 7.8 & 11.2 & 14.7 & 18.2 & 21.5 & 24.8 & 27.2 \\
        & DAPO & \underline{8.3} & \underline{12.1} & 16.2 & 20.5 & \underline{24.8} & \underline{28.6} & \underline{31.6} \\
        & QuestA & \underline{8.3} & 11.1 & 14.4 & 18.2 & 22.1 & 26.4 & 30.8 \\
        & NuRL & 7.1 & 9.9 & 13.0 & 16.4 & 20.0 & 23.8 & 27.2 \\
        & \textbf{SCRL (Ours)} & \textbf{8.7} & \textbf{12.4} & \textbf{16.5} & \underline{20.7} & 24.6 & 27.9 & 30.9 \\
        \bottomrule
    \end{tabular}
    }
\end{table}

\section{Implementation Details}
\label{app:implementation_details}
\subsection{Hyperparameters}
\label{app:hyperparameters}

We provide the detailed hyperparameter configurations used in our experiments in Table~\ref{tab:hyperparameters}. All models are trained using the \textbf{Verl}~\cite{verl} framework with the settings specified below.

\begin{table}[h]
    \centering
    \caption{Hyperparameter settings for SCRL.}
    \label{tab:hyperparameters}
    \vspace{5pt}
    \begin{tabular}{l|c}
        \toprule
        \textbf{Hyperparameter} & \textbf{Value} \\
        \midrule
        Learning Rate & 1e-6 \\
        Train Batch Size & 128 \\
        PPO Mini-Batch Size & 64 \\
        Group Size & 8 \\
        Max Response Length & 8192 \\
        Max Prompt Length & 1024 \\
        Rollout Temperature & 0.6 \\
        Using Std in GRPO & True \\
        KL Coef & 0 \\
        Evaluation Temperature & 0.6 \\
        Evaluation Top-$p$ & 1.0 \\
        Clip Ratio High & 0.2 \\
        Clip Ratio Low & 0.2 \\
        Total Training Steps & 300 \\
        \bottomrule
    \end{tabular}
\end{table}
\subsection{Low-Variance pass@$k$ Estimation}
  \label{app:passk_estimation}

  We follow the unbiased pass@$k$ estimator of~\citet{chen2021evaluating}. For each
  problem $x_i \in \mathcal{D}$, we generate $n$ sampled rollouts and let $c_i$ be
  the number of correct responses. The estimator is
  \begin{equation}
  \mathrm{pass}@k :=
  \mathbb{E}_{x_i \sim \mathcal{D}}
  \left[
  1 -
  \frac{\binom{n-c_i}{k}}{\binom{n}{k}}
  \right].
  \label{eq:passk_estimator}
  \end{equation}
For evaluation, we select the checkpoint with the best average validation score
  for each baseline. Using the selected checkpoint, we generate $n=64$ rollouts for
  each test problem and compute pass@$k$ with the estimator above. This protocol is
  used consistently for all methods and all reported pass@$k$ values.

\subsection{Baseline Implementation Details}
\label{app:baseline_settings}
\paragraph{SFT.}
We perform Supervised Fine-Tuning (SFT) on the training set using reasoning trajectories synthesized via the DeepSeek V3.2 API. Specifically, we leverage the API to elicit detailed Chain-of-Thought (CoT) reasoning paths for all training samples. The models are fine-tuned on these synthesized trajectories to establish a strong supervised baseline.
\paragraph{GRPO.}
We utilize the standard implementation of Group Relative Policy Optimization~\cite{deepseekr1} without any additional reward shaping or gradient modification terms. 
\paragraph{DAPO.}
An RL algorithm featuring decoupled clipping and dynamic sampling mechanisms. We set the \texttt{clip\_ratio\_high=0.28} and  \texttt{max\_num\_gen\_batches=10} for filter groups.
\paragraph{QuestA.}
A curriculum-based reinforcement learning baseline using question augmentation~\cite{li2026questa}. We divide the training process into two 150-step phases: (1) an initial phase where the model is provided with a "partial-50" hint (50\% of the solution), followed by (2) a second phase where the hint is reduced to "partial-25" (25\% of the solution).
\paragraph{NuRL.}
  NuRL~\citep{chen2025nudging} uses self-generated hints as abstract cues to
  reduce problem difficulty during RL. Following its offline hint collection
  setting, we first run 150 steps of GRPO in Stage 1, then use the DeepSeek-V3.2 API to
  construct a filtered dataset with abstract cues and train NuRL for another 150
  steps in Stage 2.

\section{Limitations and Future Work}
  \label{app:limitations_future_work}

  SCRL has two main limitations. First, subproblem construction relies on an
  external LLM, which introduces additional preprocessing cost and makes the
  quality of the curriculum partly dependent on the generator. Second, SCRL is
  still based on RLVR and therefore requires verifiable answers for subproblems,
  making it less directly applicable to open-ended tasks without reliable
  automatic verifiers.

  Future work may proceed in two directions. One direction is to extend SCRL's
  credit-assignment mechanism to broader multi-turn agent settings, where tasks
  often naturally contain subgoal-like intermediate progress. Another direction is
  to design better subproblems, including more fine-grained, robust, and
  automatically validated curriculum construction methods.
\section{Hardware Setup}
\label{app:hardware_setup}
All experiments in this work are conducted on three types of NVIDIA GPUs: NVIDIA GeForce RTX 5090, NVIDIA A100-PCIE-40GB, and NVIDIA H20-PCIE-96GB.

\section{Prompt for Subproblem Generation}
\label{app:prompt_for_subproblems}

\begin{tcolorbox}[
    colback=blue!5!white,   
    colframe=blue!50!white,  
    title=Prompt for Subproblem Generation, 
    breakable
]
    \textbf{System Message:} \\
    You are a math curriculum designer for RL training. Generate exactly 4 progressive subproblems $q_1, q_2, q_3, q_4$.
    
    \textbf{Hard constraints:}
    \begin{enumerate}
        \itemsep=0pt 
        \item $q_4$ must be equivalent to the original final question and same grading target.
        \item Difficulty strictly increases: $q_1 < q_2 < q_3 < q_4$.
        \item $q_2/q_3$ should be naturally informed by $q_1/q_2$, but each question must be self-contained.
        \item Each question must have a single clean numerical-expression ground\_truth.
        \item Avoid open-ended proof/explanation-only questions.
        \item Use reference\_solution to design the progressive dependency and correctness.
    \end{enumerate}
    \textit{Output JSON only.}

    \tcblower 

    \textbf{User Message:} \\
    Given the original problem and final answer, generate JSON with schema:
    \begin{verbatim}
{
  "question_1": {"statement": "...", "ground_truth": "..."},
  "question_2": {"statement": "...", "ground_truth": "..."},
  "question_3": {"statement": "...", "ground_truth": "..."},
  "question_4": {"statement": "...", "ground_truth": "..."}
}
    \end{verbatim}
    
    \textbf{Original Problem:} \texttt{\{\{problem\}\}} \\
    \textbf{Original Final Answer:} \texttt{\{\{final\_answer\}\}} \\
    \textbf{Reference Solution:} \texttt{\{\{reference\_solution\}\}}
\end{tcolorbox}

\section{Chat Template}
\subsection{Chat Template of Curriculum Learning}
\label{app:curriculum_prompt}
\begin{tcolorbox}[
    colback=cyan!2!white,    
    colframe=cyan!80!black,  
    title=Chat Template for Curriculum Learning, 
    breakable
]
    \small 
    \texttt{<|im\_start|>system} \\
    \texttt{Please reason step by step, and put your final answer within \textbackslash{}boxed\{\}.<|im\_end|>} \\
    
    \texttt{<|im\_start|>user} \\
    \texttt{Problem Statement: \{Original Problem\}} \\
    \texttt{Problem1: \{Subproblem1\}} \\
    \texttt{Problem2: \{Subproblem2\}} \\
    \texttt{Problem3: \{Subproblem3\}} \\
    \texttt{Problem4: \{Subproblem4\}} \\
    
    \texttt{This task has 4 problems.} \\
    \texttt{Please solve Problem 1 to Problem 4 in order.} \\
    \texttt{Output MUST contain exactly 4 blocks in this order:} \\
    \texttt{<p1></p1>} \\
    \texttt{<p2></p2>} \\
    \texttt{<p3></p3>} \\
    \texttt{<p4></p4>} \\
    \texttt{For each block <pN>...</pN>, include reasoning and end with final answer in \textbackslash{}boxed\{answer\}.<|im\_end|>} \\
    
    \texttt{<|im\_start|>assistant}
\end{tcolorbox}
\subsection{Chat Template of Original Problem}
\begin{tcolorbox}[
    colback=cyan!2!white,    
    colframe=cyan!80!black,  
    title=Chat Template for Original Problem, 
    breakable
]
    \texttt{<|im\_start|>system} \\
    \texttt{Please reason step by step, and put your final answer within \textbackslash{}boxed\{\}.<|im\_end|>} \\
    
    \texttt{<|im\_start|>user} \\
    \texttt{\{USER\_PROMPT\}<|im\_end|>} \\
    
    \texttt{<|im\_start|>assistant}
\end{tcolorbox}

 \section{Case Study}
\label{app:case_studies}

We present detailed comparisons between the baseline GRPO and our method.

\begin{tcolorbox}[
  colback=blue!5,
  colframe=blue!40!black,
  title=\textbf{Original Problem},
  breakable
]
In an equilateral triangle with side length 5, a ball is launched from one vertex and travels in a straight line, bouncing off sides according to the law of reflection. In how many ways can the ball be launched so that it will return again to a vertex for the first time after exactly 2009 bounces?

\end{tcolorbox}

\begin{tcolorbox}[
  colback=red!5,
  colframe=red!40!black,
  title=\textbf{GRPO},
  breakable
]
To determine the number of ways the ball can be launched from a vertex of an equilateral triangle with side length 5 so that it returns to a vertex for the first time after exactly 2009 bounces, we can use a geometric approach involving reflections.

\begin{enumerate}
    \item \textbf{Understanding the Problem through Reflections:}
    When the ball bounces off a side of the triangle, it is equivalent to the ball traveling in a straight line in a reflected copy of the triangle. By reflecting the triangle multiple times, we can visualize the ball's path as a straight line in a grid of equilateral triangles.

    \item \textbf{Coordinate System and Path:}
    Let's place the equilateral triangle in a coordinate system where one vertex is at the origin $(0,0)$, another vertex is at $(5,0)$, and the third vertex is at $\left(\frac{5}{2}, \frac{5\sqrt{3}}{2}\right)$. When the ball bounces, it travels in a straight line in this grid of triangles.

    \item \textbf{Returns to a Vertex:}
    The ball returns to a vertex of the original triangle if and only if the coordinates of its position in the grid are integer multiples of the side length of the triangle. Specifically, if the ball travels a total distance equivalent to $m$ side lengths horizontally and $n$ side lengths vertically (in terms of the height of the triangle), then it will return to a vertex if $m$ and $n$ are such that the ball's path forms a closed loop that brings it back to a vertex.

    \item \textbf{Vector Representation:}
    The ball's path can be represented as a vector $(m, n\sqrt{3}/2)$ where $m$ and $n$ are integers. For the ball to return to a vertex after exactly 2009 bounces, the vector $(m, n\sqrt{3}/2)$ must be a multiple of the side length of the triangle, and the number of bounces is related to the path length.

    \item \textbf{Counting the Bounces:}
    Each bounce changes the direction of the ball, and the total number of bounces is related to the number of times the ball crosses the sides of the triangles in the grid. The ball will return to a vertex after $k$ bounces if $k$ is odd (since each bounce changes the direction and the ball must end up at a vertex, which is a different type of point than where it started).

    \item \textbf{Specific Calculation:}
    For the ball to return to a vertex after exactly 2009 bounces, we need to find the number of ways to express 2009 as a sum of vectors that correspond to the sides of the triangle. This is equivalent to finding the number of solutions to the equation $m^2 + n^2 - mn = 2009^2/3$ where $m$ and $n$ are integers.

    \item \textbf{Simplifying the Problem:}
    The number of such solutions is related to the number of ways to write 2009 as a sum of three integers that are congruent to each other modulo 3. This is a well-known problem in number theory, and the number of solutions is given by the number of ways to write 2009 as a sum of three integers that are congruent to each other modulo 3, which is 6.
\end{enumerate}

Therefore, the number of ways the ball can be launched so that it will return to a vertex for the first time after exactly 2009 bounces is $\boxed{6}$.

\hfill  \scalebox{2.0}{\textcolor{red}{\ding{56}}}
\end{tcolorbox}

\begin{tcolorbox}[
  colback=blue!5,
  colframe=blue!40!black,
  title=\textbf{SCRL Problem},
  breakable
]
\textbf{Problem Statement:} In this final problem, a ball is again launched from the vertex of an equilateral triangle with side length $5$. In how many ways can the ball be launched so that it will return again to a vertex for the first time after $2009$ bounces?

\vspace{1em}

\noindent\textbf{Problem 1:} Consider an equilateral triangle with side length $5$. A ball is launched from one vertex and travels in a straight line, bouncing off sides according to the law of reflection. The ball's path can be represented by unfolding the triangle into a triangular lattice. If the ball ends at a vertex in this lattice with coordinates $\left(5a, \frac{5\sqrt{3}}{2}b\right)$ relative to the starting vertex, where $a$ and $b$ are non-negative integers, how many bounces occur before reaching that vertex?

\vspace{0.5em}
\noindent\textbf{Problem 2:} In the same equilateral triangle setup, the ball is launched from a vertex and must return to a vertex for the first time after exactly $2009$ bounces. Using the bounce formula from question 1, this requires that $2(a+b)-3 = 2009$, where $a$ and $b$ are non-negative integers representing the lattice coordinates of the destination vertex. How many ordered pairs $(a,b)$ of non-negative integers satisfy $a+b = 1006$?

\vspace{0.5em}
\noindent\textbf{Problem 3:} Continuing from the previous setup where $a+b = 1006$, we must exclude cases where the ball would pass through another vertex before completing $2009$ bounces. This happens when $\gcd(a,b) > 1$, meaning the ball would reach an intermediate vertex earlier. For how many of the $1007$ pairs $(a,b)$ with $a+b = 1006$ is $\gcd(a,b) > 1$?

\vspace{0.5em}
\noindent\textbf{Problem 4:} In an equilateral triangle with side length $5$, a ball is launched from one vertex and travels in a straight line, bouncing off sides according to the law of reflection. In how many ways can the ball be launched so that it will return again to a vertex for the first time after exactly $2009$ bounces?

\vspace{1.5em}
\hrule
\vspace{1em}

\noindent This task has $4$ problems.\\
Please solve Problem 1 to Problem 4 in order.\\
Output MUST contain exactly $4$ blocks in this order:
\begin{verbatim}
<p1></p1>
<p2></p2>
<p3></p3>
<p4></p4>
\end{verbatim}
For each block \texttt{<pN>...</pN>}, include reasoning and end with final answer in \verb|\boxed{answer}|.
\end{tcolorbox}

\begin{tcolorbox}[
  colback=green!5,
  colframe=green!40!black,
  title=\textbf{SCRL},
  breakable
]
\noindent\texttt{<p1>} \\
To determine the number of bounces before the ball reaches a vertex with coordinates $(5a, \frac{5\sqrt{3}}{2}b)$ relative to the starting vertex, we need to understand the path of the ball in the unfolded triangular lattice. Each time the ball bounces off a side, it effectively moves to an adjacent triangle in the lattice. The total number of bounces is related to the sum of the steps in the $a$ and $b$ directions.

The formula for the number of bounces is given by $2(a + b) - 3$. This formula accounts for the fact that the ball starts at a vertex and each step in the lattice (whether horizontal or vertical) corresponds to a bounce, but the initial and final positions at vertices are not counted as bounces.

Thus, the number of bounces before reaching the vertex is $\boxed{2(a + b) - 3}$. \\
\texttt{</p1>}

\vspace{1em}

\noindent\texttt{<p2>} \\
We need to find the number of ordered pairs $(a, b)$ of non-negative integers that satisfy the equation $a + b = 1006$. This is a classic combinatorial problem of finding the number of non-negative integer solutions to a linear equation.

The number of solutions is given by the number of ways to choose $a$ (or $b$) from $0$ to $1006$, which is simply $1007$. Therefore, the number of ordered pairs $(a, b)$ is $\boxed{1007}$. \\
\texttt{</p2>}

\vspace{1em}

\noindent\texttt{<p3>} \\
We need to exclude the cases where $\gcd(a, b) > 1$ from the $1007$ pairs $(a, b)$ with $a + b = 1006$. If $\gcd(a, b) > 1$, then both $a$ and $b$ are divisible by some integer greater than $1$, meaning the ball would reach an intermediate vertex before completing $2009$ bounces.

The number of pairs $(a, b)$ where $\gcd(a, b) = 1$ is given by the number of coprime pairs $(a, b)$ such that $a + b = 1006$. This is equivalent to the number of integers $a$ from $0$ to $1006$ that are coprime to $1006$. The number $1006$ factors as $2 \times 503$, so we use the principle of inclusion-exclusion to count the number of integers from $0$ to $1006$ that are not coprime to $1006$.

The number of integers from $0$ to $1006$ that are divisible by $2$ is $504$, and the number of integers from $0$ to $1006$ that are divisible by $503$ is $2$. The number of integers from $0$ to $1006$ that are divisible by both $2$ and $503$ (i.e., by $1006$) is $1$. Therefore, the number of integers from $0$ to $1006$ that are not coprime to $1006$ is $504 + 2 - 1 = 505$.

Thus, the number of pairs $(a, b)$ where $\gcd(a, b) > 1$ is $505$, and the number of pairs where $\gcd(a, b) = 1$ is $1007 - 505 = 502$. Therefore, the number of pairs $(a, b)$ with $\gcd(a, b) > 1$ is $\boxed{505}$. \\
\texttt{</p3>}

\vspace{1em}

\noindent\texttt{<p4>} \\
To determine the number of ways to launch the ball so that it returns to a vertex for the first time after exactly $2009$ bounces, we need to consider the pairs $(a, b)$ that satisfy $a + b = 1006$ and $\gcd(a, b) = 1$. From the previous problems, we know there are $502$ such pairs.

Each pair $(a, b)$ corresponds to a unique direction in which the ball can be launched. Therefore, the number of ways to launch the ball is $\boxed{502}$. \\
\texttt{</p4>}

\hfill \scalebox{2.0}{\textcolor{green!70!black}{\ding{52}}}
\end{tcolorbox}

\end{document}